\documentclass[]{interact}

\usepackage{epstopdf}
\usepackage[caption=false]{subfig}

\usepackage[numbers,sort&compress]{natbib}
\bibpunct[, ]{[}{]}{,}{n}{,}{,}
\renewcommand\bibfont{\fontsize{10}{12}\selectfont}

\theoremstyle{plain}
\newtheorem{theorem}{Theorem}[section]
\newtheorem{lemma}[theorem]{Lemma}

\theoremstyle{definition}
\newtheorem{definition}[theorem]{Definition}
\newtheorem{example}[theorem]{Example}

\theoremstyle{remark}

\usepackage{graphicx}
\usepackage{enumitem}
\usepackage{parskip}
\usepackage{fancybox}
\usepackage{latexsym}
\usepackage{amsmath}
\usepackage{amssymb}
\usepackage{xspace}
\usepackage{float}
\usepackage{graphicx}
\usepackage{pstricks}
\usepackage{psfrag}
\usepackage{boxedminipage}
\usepackage{changepage}
\usepackage{graphicx}
\usepackage{color}
\usepackage{psfrag}
\usepackage{yhmath}
\usepackage{array, calc}
\usepackage{rotating}
\usepackage{hyperref}
\usepackage{multirow, subfig}
\usepackage{proof}
\usepackage{amsbsy}

\newcommand{\comment}[1]{}

\newcommand{\dec}           {D}
\newcommand{\nats}          {\mathbb{N}}
\newcommand{\ints}          {\mathbb{Z}}

\newcommand{\Gurobi}         {\textsf{Gurobi}\xspace}
\newcommand{\CPLEX}          {\textsf{CPLEX}\xspace}
\newcommand{\SCIP}           {\textsf{SCIP}\xspace}
\newcommand{\GLPK}           {\textsf{GLPK}\xspace}
\newcommand{\CutSat}         {\textsf{CutSat}\xspace}
\newcommand{\IntSat}         {\textsf{IntSat}\xspace}
\newcommand{\MiniSAT}        {\textsf{MiniSAT}\xspace}

\newcommand{\proc}{\textsf{Conflict\_}\textsf{analysis\_}\textsf{Backjump\_Learn}}
\newcommand{\mio}{\lor}
\newcommand{\mios}{\lor\ldots\lor}

\newcommand{\no}[1]{\overline{#1}}
\newcommand{\snp}          {$\mbox{}$\hspace{6pt}}
\newcommand{\bleq}          {\!\preceq\!}
\newcommand{\cleq}          {  \preceq  }
\newcommand{\cgeq}          {  \succeq  }
\newcommand{\ceq}          {  =  }
\newcommand{\height}       {\textrm{height}}

\renewcommand{\b}[1]{{\textbf{#1}}}

\newcommand{\warning}[1]{{#1}}

\begin{document}

\articletype{ARTICLE}

\title{\IntSat: Integer Linear Programming by Conflict-Driven Constraint-Learning}

\author{
  \name{
    Robert Nieuwenhuis\textsuperscript{a} and Albert Oliveras\textsuperscript{a} and Enric Rodr\'\i guez-Carbonell\textsuperscript{a}\thanks{CONTACT Enric Rodr\'\i guez-Carbonell. Email: erodri@cs.upc.edu}
  }
  \affil{
    Barcelogic and Department of Computer Science, Universitat Polit\`ecnica de Catalunya, Barcelona, Spain
  }
}

\maketitle

\begin{abstract}
  State-of-the-art SAT solvers are nowadays able to handle huge
  real-world instances. The key to this success is the so-called
  \emph{Conflict-Driven Clause-Learning (CDCL)} scheme, which
  encompasses a number of techniques that exploit the conflicts that
  are encountered during the search for a solution. In this article we
  extend these techniques to Integer Linear Programming (ILP), where
  variables may take general integer values instead of purely binary
  ones, constraints are more expressive than just propositional
  clauses, and there may be an objective function to optimise. We
  explain how these methods can be implemented efficiently, and
  discuss possible improvements. Our work is backed with a basic
  implementation that shows that, even in this far less mature stage,
  our techniques are already a useful complement to the state of the
  art in ILP solving.
\end{abstract}


\begin{amscode}
90C10  
\end{amscode}

\section{Introduction}
\label{intro}

{Since the early days of computer science, propositional logic
  has been recognised as one of its cornerstones. A fundamental result
  in the theory of computing is the proof by Cook that SAT, that is,
  the problem of deciding whether a propositional formula is
  satisfiable or not, is NP-complete \cite{Cook1971STOC}. It was soon
  realised that, in consonance with this fact, a wide range of
  combinatorial problems could be expressed in SAT
  \cite{Kar72}. Hence, due to its potential practical implications,
  since then an extensive research has been underway of how SAT could
  be solved in an automated and efficient way
  \cite{HandbookOfSAT2009}. As a result of this work, particularly
  intensive in the last two decades
  \cite{Gomes:1998:BCS:295240.295710,MarquesSakallah1999IEEE,Zhangetal2001ICCAD,GoldbergNovikov2002DATE,BeameKS04,EenBiere2005SAT,Huang2007,Pipatsrisawat2010JAR,DBLP:conf/cade/JarvisaloHB12,DBLP:journals/jair/HeuleJLSB15},
  now SAT solvers routinely handle formulas coming from real-world
  applications with hundreds of thousands of variables and millions of
  clauses.}

{ State-of-the-art SAT solvers are essentially based on the
  \emph{Davis-Putnam-Logemann-Loveland (DPLL)} procedure
  \cite{DavisPutnam1960,Davisetal1962CACM}. In a nutshell, DPLL is a
  backtracking algorithm that searches for a (feasible) solution by
  intelligently traversing the search space. At each step a
  \emph{decision} is made: a variable is selected for branching and is
  assigned a value either 0 or 1. Then the consequences of that
  decision are \emph{propagated}, and variables that are forced to a
  value are detected. Each time a falsified clause (i.e., a
  \emph{conflict}) is identified, backtracking is executed.
  \warning{Backtracking consists in undoing all assignments up to
  the last decision and forcing the}
  branching variable to take the other value. When both values of the
  branching variable have already been tried without success, then the
  previous decision is backtracked. If the decision is the first one,
  and therefore there is no previous decision, then the formula can be
  declared unsatisfiable, i.e, infeasible.}

{This simple description is, however, far from the current
  implementations of SAT solvers. What accounts for their
  success is the so-called \emph{Conflict-Driven Clause-Learning
    (CDCL)} scheme, which enhances DPLL with a number of techniques:
  \begin{itemize}
  \item \emph{conflict analysis} and \emph{backjumping} (i.e.,
    non-chronological backtracking), which improves (chronological)
    backtracking \cite{MarquesSakallah1999IEEE};
  \item \emph{learning} (that is, addition) of new clauses generated from
    conflicts \cite{DECHTER1990273};
  \item variable decision heuristics that select the most active
    variables in recent conflicts, like the \emph{VSIDS} heuristic
    \cite{Chaff2001};
  \item value decision heuristics that select promising values for the
    chosen decision variable, such as the \emph{last phase} strategy
    \cite{PipatsrisawatD153};
  \item data structures such as the \emph{2-watched literal scheme}
    \cite{Chaff2001} for efficiently identifying propagations and
    conflicts;
  \item \emph{restarts} \cite{10.1007/BFb0017434};
%
%
  \item \emph{clause cleanups} that periodically delete the
    least useful learnt clauses, e.g. based on their activity in
    conflicts \cite{GoldbergNovikov2002DATE}.
  \end{itemize}
}

{Since the problems of Integer Linear Programming (ILP) and
  SAT are both NP-complete, they reduce one another. Thus a natural
  question is how solvers of the two areas compare. It turns out that,
  in spite of the maturity of the ILP solving technology and its
  astonishing achievements \cite{Bixby2017},
  %
  %
  for problem instances of a more combinatorial (as opposed to
  numerical) sort, SAT solvers with an encoding into propositional
  clauses can outperform even the best commercial ILP solvers run on
  a --far more compact-- ILP formulation \cite{DBLP:journals/jfi/Aloul06,DBLP:conf/alcob/BrownZG20}.

  The different nature of the algorithms underlying ILP solvers (based
  on branch-and-cut and the simplex method) and SAT solvers motivate
  this work, in which we aim at pushing the techniques of CDCL beyond SAT,
  while handling ILP constraints natively at all levels.}

{
  To do so, the following issues need to be addressed:
  \begin{itemize}
  \item variables are no longer binary, and may take general integer
    values. Moreover, the domains of values may be bounded (the
    variable can only take a value within an interval) or unbounded
    (the variable may take any value in $\ints$). Should we do case
    analysis by taking concrete values of variables or based on lower
    and upper \emph{bounds}?
    Also,
    the notion of decision has to be determined: does a decision fix a
    variable to a value, or does it split its domain? The value
    decision heuristics have to be defined as well. Finally, unbounded domains
    pose a problem with the termination of the search algorithm.
  \item constraints are not just clauses any more, and can be general
    linear inequalities. Hence the propagation mechanism
    (be it of variable values or of bounds) 
    has to be
    redefined, as well as the algorithms and data structures for
    efficiently detecting when a propagation can be triggered or a
    conflict has arisen. Unlike in SAT, attention has to be paid to
    arithmetic and numerical problems, so as to ensure the soundness
    of the procedures. Most importantly, conflict analysis has to be
    generalised in such a way that backjumping and learning are
    possible.
  \item problems are no longer purely feasibility checks, and may
    require to optimise an objective function. In addition to enabling
    the search algorithm to optimise, for practical reasons the
    information provided by the objective function has to be
    integrated into the decision heuristics.
  \end{itemize}
}

\warning{ Building on top of the early ideas in
  \cite{Nieuwenhuis2014CP}, here we present the \IntSat method, a
  family of new different conflict-driven \emph{constraint}-learning
  algorithms for ILP. We illustrate the method with two algorithms,
  which differ in the way in which the aforementioned issues on
  extending conflict analysis and learning are resolved.}
We provide detailed explanations of how these algorithms can be implemented efficiently and also report an extensive up-to-date experimental analysis, comparing a basic implementation of \IntSat against the best commercial (simplex-based) ILP solvers. The results show that, even in this far less mature stage, our techniques already are a useful complement to the state of the art in ILP solving, finding (good) solutions faster in a significant number of instances, especially in those of a more combinatorial (as opposed to numerical) nature.


  { This paper is structured as follows\footnote{
\textbf{Note to the reviewers:} the proofs of the appendix could be left out of the final published version (unless you suggest otherwise).
  }. Preliminary background
  on SAT and ILP is reviewed in Section~\ref{preliminaries}. Section
  \ref{from-sat-to-ilp} introduces \warning{our two \IntSat} algorithms that generalise CDCL
  from SAT to ILP. After presenting the main hindrance in this
  generalisation (Section \ref{rounding-problem}) and reviewing basic
  properties of propagation in ILP (Section \ref{bound-propagation}),
  the common part of these algorithms is exposed (Section
  \ref{intsat}). Then their differences are described in detail
  (Sections \ref{set-based} and \ref{cut-based}, respectively). A
  discussion on extensions (Section \ref{extensions}) concludes
  Section \ref{from-sat-to-ilp}. Section \ref{implem} is devoted to
  implementation issues, while in Section \ref{experiments} the
  results of an experimental evaluation are reported. In Section
  \ref{further}, ideas for future research are outlined. Finally,
  Section \ref{conc} completes this article with an account of related
  work and conclusions.}


\section{Preliminaries}
\label{preliminaries}

\subsection{Propositional Satisfiability}
\label{sat}

{Let $X$ be a finite set of \emph{propositional variables}. If
  $x \in X$, then $x$ and $\no{x}$ are \emph{literals} of $X$.
  The \emph{negation} of a literal $l$, written $\no{l}$,
  denotes $\no{x}$ if $l$ is $x$, and $x$ if $l$ is
  $\no{x}$. A \emph{clause} is a disjunction of literals
  $l_1 \lor\ldots\lor l_n$. A (CNF) \emph{formula} is a conjunction of
  one or more clauses $C_1 \land\ldots\land C_n$. When it leads to no
  ambiguities, we will sometimes consider a clause as the set of its
  literals, and a formula as the set of its clauses. }

{A (partial truth) \emph{assignment} $A$ is a set of literals
  such that $\{ x, \no{x} \} \subseteq A$ for no $x$. A literal
  $l$ is \emph{true} in $A$ if $l \in M$, is \emph{false} in $A$ if
  $\no{l} \in A$, and is \emph{undefined} in $A$ otherwise. A
  clause $C$ is true in $A$ if at least one of its literals is true in
  $A$, is false or a \emph{conflict} if all of its literals are false
  in $A$, and is undefined otherwise. A formula $F$ is true in $A$ if
  all of its clauses are true in $A$. In that case, we say that $A$ is
  a \emph{solution} to $F$, and that $F$ is \emph{satisfied} by $A$.
  The problem of \emph{SAT} consists in deciding, given a formula $F$,
  whether $F$ is satisfiable, that is, there exists a solution to $F$.
  The systems that solve SAT are called \emph{SAT solvers}.}

{The core of a \emph{Conflict-Driven Clause-Learning (CDCL)}
  SAT solver is described} in the following algorithm, where $A$ is
seen as an (initially empty) stack: 

\begin{enumerate}[label=\arabic*.]
\item
{\textsf{Propagate}: while possible and no conflict appears,
if, for some clause $l\mio C$, \ $C$ is false in $A$ and $l$ is undefined,
push $l$ onto $A$, associating to $l$ the \emph{reason clause} $C$.}
\item
\textbf{if} there is no conflict\\
  \snp \textbf{if} all variables are defined in $A$, output `solution $A$' and halt.\\
  \snp \textbf{else} {\textsf{Decide}: push an undefined literal $l$, marked
  as a \emph{decision}, and go to step 1.}
\item {\textbf{if} there is a conflict and $A$ contains no decisions,}\\
  \snp output `unsatisfiable' and halt.
\item \label{label}
  {\textbf{if} there is a conflict and $A$ contains some decision,} \\
  \snp use a clause data structure $C$, {the \emph{conflicting clause}.}
  Initially, let $C$ be any conflict.
\begin{enumerate}[label=\theenumi\arabic*.]
\item
{\textsf{Conflict analysis}}:

\underline{Invariant:} $C$ is false in $A$, that is, if $l\in C$ then $\no{l}\in A$.

{Let $l$ be the literal of $C$ whose negation is topmost in $A$,
and $D$ be the reason clause of $\no{l}$.
Replace $C$ by $(C\setminus \{l\}) \lor D$.}
Repeat this until there is only one literal $l_{top}$ in $C$
such that $\no{l_{top}}$ is, or is above, $A$'s topmost decision.
\item
{\textsf{Backjump}}: pop literals from $A$
until either there are no decisions in $A$ or,
for some $l$ in $C$ with $l\not=l_{top}$, there are no decisions
above $\no{l}$ in $A$.
\item
{\textsf{Learn}}: add the final $C$ as a new clause, and go to 1 (where $C$ propagates $l_{top}$).
\end{enumerate}
\end{enumerate}

{The literal $l_{top}$ at step 4 is called the \emph{first
    implication point (1UIP)}, whence the conflict analysis above is said to
  follow a \emph{1UIP scheme} \cite{Zhangetal2001ICCAD}.}

{Implicitly in this description of the algorithm, conflict
  analysis uses \emph{resolution} \cite{Robinson1965JACM} to inspect
  back the cause of a conflict. Given a variable $x$ and two clauses
  of the form $x \lor A$ and $\no{x} \lor B$ (the \emph{premises}),
  the resolution rule infers a new clause $A \lor B$ (the
  \emph{resolvent}). Graphically:}

{
    $$
    \infer{A \lor B}
    { {x} \lor A & \no{{x}} \lor B}
    $$
    }
    {Back to the algorithm, when the current conflicting clause
      $C$ is replaced by $(C\setminus \{l\}) \lor D$, where
      $l$ is the literal of $C$ whose negation is topmost in $A$ and
      $D$ is the reason clause of $\no{l}$, in fact resolution is
      being applied between $C$ (viewed as
      $l \lor (C\setminus \{l\})$) and $\no l \mio D$.}

\begin{example}

  { Let us assume we have clauses $\no x_{1} \lor x_{2}$,
    $\no x_{3} \lor x_{4}$, $\no x_{5} \lor x_{6}$,
    $\no x_{5} \lor \no x_{6} \lor x_{7}$ and
    $\no x_{2} \lor \no x_{5} \lor \no x_{7}$. Since there is nothing
    to propagate at the beginning, we decide to make $x_{1}$ true and
    push it onto the stack of the assignment. Now due to
    $\no x_{1} \lor x_{2}$, literal $x_{2}$ is propagated and pushed
    onto the stack with reason clause $\no x_{1}$. Again there is
    nothing to propagate, so we decide to make $x_{3}$ true. Due to
    $\no x_{3} \lor x_{4}$, literal $x_{4}$ is propagated and pushed
    onto the stack with reason clause $\no x_{3}$. Yet again there is
    nothing else to propagate, so we decide to make $x_{5}$ true. Due
    to $\no x_{5} \lor x_{6}$, literal $x_{6}$ is propagated and
    pushed onto the stack with reason clause $\no x_{5}$. And in turn,
    due to $\no x_{5} \lor \no x_{6} \lor x_{7}$, literal $x_{7}$ is
    propagated and pushed onto the stack with reason clause
    $\no x_{5} \lor \no x_{6}$. Now clause
    $\no x_{2} \lor \no x_{5} \lor \no x_{7}$ has become false under
    the current assignment
    $({x_{1}}^{\mathsf{d}}, x_{2}, {x_{3}}^{\mathsf{d}}, x_{4},
    {x_{5}}^{\mathsf{d}}, x_{6}, x_{7})$, where decisions are marked with a
    superscript $\textsf{d}$.}

  \begin{itemize}

  \item {{\textsf{Conflict analysis}}: The conflicting clause $C$ is
      initially $\no x_{2} \lor \no x_{5} \lor \no x_{7}$. In the
      first iteration of conflict analysis, we replace $\no x_{7}$ in
      $C$ by the reason clause $\no x_{5} \lor \no x_{6}$ of $x_{7}$,
      which yields $\no x_{2} \lor \no x_{5} \lor \no x_{6}$ as the
      new $C$. In the second iteration we similarly replace
      $\no x_{6}$ in $C$, now
      $\no x_{2} \lor \no x_{5} \lor \no x_{6}$, with the reason
      clause $\no x_{5}$ of $x_{6}$, which gives
      $\no x_{2} \lor \no x_{5}$. Alternatively, since as explained
      above for each propagated literal $l$ with reason clause $D$ we
      are applying resolution between $l \mio D$ and $C$, the sequence
      of these two steps can be represented graphically as follows:
      $$
      \infer{\no x_{2} \lor \no x_{5}}
      {
        \boldsymbol{x_{6}} \lor \no x_{5}
        &
        \infer{\no x_{2} \lor \no x_{5} \lor \no{\boldsymbol{x_{6}}}}{
          \boldsymbol{x_{7}} \lor \no x_{5} \lor \no x_{6} & \no x_{2} \lor \no x_{5} \lor \no{\boldsymbol{x_{7}}}}
      }
      $$
      where the variable that is eliminated at each step is
      highlighted in bold face, and the clauses on the right are the
      conflicting clauses, namely $\no x_{2} \lor \no x_{5} \lor \no{x_{7}}$,
      $\no x_{2} \lor \no x_{5} \lor \no{x_{6}}$ and
      $\no x_{2} \lor \no x_{5}$. Note that each of
      these clauses is false under the current assignment, as ensured
      by the invariant.
    }

    {Now there is a single literal $l_{top}$ in $C$ (literal
      $\no x_{5}$) such that $\no{l_{top}}$ is, or is above, the stack's
      topmost decision (in this case, $x_{5}$). Therefore we backjump.}

  \item {{\textsf{Backjump}}: We pop literals from the stack until
      either (i) there are no decisions in the stack, or (ii) for some
      $l$ in $C$ with $l\not=l_{top}$, there are no decisions above
      $\no{l}$ in the stack. In this case (ii) applies with literal
      $\no x_{2}$ as $l$, and so we pop the literals from $x_{7}$ down
      to $x_{3}$, leaving the stack as
      $({x_{1}}^{\mathsf{d}}, x_{2})$. Note in particular that the
      decision ${x_{3}}$ and its propagation $x_{4}$, which are
      irrelevant to the conflict, are \emph{jumped} over. The
      intuition is that, if we had had the clause
      $\no x_{2} \lor \no x_{5}$ in the clause database at the time we
      decided ${x_{3}}$, we would have propagated $\no x_{5}$ before
      any decision. Precisely, the following \emph{Learn} step will allow making
      this propagation.}

  \item {{\textsf{Learn}}: The final $C$, namely
      $\no x_{2} \lor \no x_{5}$, is learned and added to the clause database.}

  \end{itemize}

\noindent
{Back in step 1, we can finally propagate $\no x_{5}$ with the
  learned clause, leading to the assignment
  $({x_{1}}^{\mathsf{d}}, x_{2}, \no x_{5})$. Now we decide $x_{4}$,
  and all clauses are satisfied. The remaining variables can now be
  decided arbitrarily, and once all of them are defined, since there
  is no conflict the algorithm terminates with a solution. \comment{\qed}}
\end{example}

\subsection{Integer Linear Programming}
\label{ilp}

{Let $X$ be a finite set of \emph{integer variables}
  $\{x_1\ldots x_n\}$. An \emph{(integer linear) constraint} over $X$
  is an expression of the form $a_1 x_1 + \cdots + a_n x_n \cleq a_0$
  where, for all $i$ in $0\ldots n$, the \emph{coefficients} $a_i$ are
  integers (some of which may be zero). Each of the terms $a_i x_i$ is
  called a \emph{monomial} of the constraint. In what follows,
  variables are always denoted by (possibly sub-indexed or primed)
  lowercase $x,y,z$, and coefficients by $a,b,c$, respectively.}

{ A constraint $a_1 x_1 +\ldots+ a_nx_n \cleq a_0$ is said to
  be \emph{normalised} if $gcd(a_1,\ldots, a_n) = 1$. For any
  constraint $a_1 x_1 +\ldots+ a_nx_n \cleq a_0$, the constraint
  $\; a_1/d\;\; x_1 \; +\ldots+\; a_n/d\;\; x_n \; \cleq \; \lfloor
  a_0/d \rfloor$, where $d = gcd(a_1,\ldots, a_n)$, is an equivalent
  normalised constraint. Hence in what follows we assume all
  constraints to be eagerly normalised.}

{An \emph{integer linear program (ILP)} over $X$ is a set $S$ of
  integer linear constraints over $X$, called the \emph{constraints}
  of the ILP, together with a linear expression of the form
  $c(x_{1}, \cdots, x_{n}) = c_1 x_1 + \cdots + c_n x_n$, called the
  \emph{objective function}. A \emph{solution} to a set of constraints
  $S$ over $X$ (and, by extension, to an ILP whose set of constraints
  is $S$) is a function $sol\colon X \to \ints$ that \emph{satisfies}
  every constraint $a_1 x_1 + \cdots + a_n x_n \cleq a_0$ in $S$, that
  is,
  $a_1 \cdot sol(x_1) + \cdots + a_n \cdot sol(x_n) \leq_\ints a_0$.
  {If a solution to $S$ exists, then $S$ (and again by
    extension, any ILP whose set of constraints is $S$) is
    \emph{feasible}, otherwise it is said to be \emph{infeasible} .}
  Without loss of generality, we will assume that the objective
  function in an ILP is to be minimised: an \emph{optimal solution} to
  an ILP with constraints $S$ and objective function $c$ is a solution
  $sol$ to $S$ such that $c(sol) \leq c(sol')$ for any solution $sol'$
  to $S$. The problem of \emph{Integer Linear Programming (ILP)}
  consists in finding an optimal solution to a given integer linear
  program. When there is no ambiguity, we will use ILP both for
  `integer linear program' as well as `integer linear
  programming'.}

{ A \emph{bound} is a one-variable constraint
  $a_1 x \cleq a_0$. Any bound can equivalently be written either as a
  \emph{lower bound} $a\bleq x$ or as an \emph{upper bound}
  $x\bleq a$. A variable $x$ is \emph{binary} in an ILP if its set of
  constraints $S$ contains the lower bound $0 \bleq x$ and the upper
  bound $x \bleq 1$, so that effectively $x$ can only take values
  either 0 or 1.}

{Given a set of constraints $S$ and a constraint $C$, we write
$S \models C$ when any solution to $S$ is also a solution to $C$. The
definition of $\models$ is lifted to sets of constraints on the
right-hand side in the natural way.}

{From constraints $C_{1}$ of the form
  $a_1 x_1 + \cdots + a_n x_n \cleq a_0$ and $C_{2}$ of the form
  $b_1 x_1 + \cdots + b_n x_n \cleq b_0$ (called the \emph{premises}),
  and natural numbers $c$ and $d$, the \emph{cut rule} derives a new
  constraint $C_{3}$ of the form
  $c_1 x_1 + \cdots + c_n x_n \cleq c_0$ (called the \emph{cut}),
  where $c_i = c a_i + d b_i$ for $i$ in $0\ldots n$. The cut rule is
  correct in the sense that it only infers consequences of the
  premises: $\{C_{1}, C_{2}\} \models C_{3}$. If for some $i > 0$ we
  have $c_i=0$, then we say that $x_i$ is \emph{eliminated} in this
  cut. Note that if $a_i b_i < 0$, then one can always choose $c$ and
  $d$ such that $x_i$ is eliminated. See
  \cite{Gomory58,Chvatal73,Schrijver86} for further discussions and
  references about Chv\'atal-Gomory cuts and their applications to
  solving ILP's.}

\begin{example}
  { Let us see an example of application of the cut rule to
    the constraints $4x + 4y + 2z \cleq 3$ and $-10x + y - z \cleq 0$.
    By multiplying the former by 5 and the latter by 2 and adding them
    up, we obtain the cut $22y + 8z \cleq 15$, in which variable $x$
    has been eliminated. Note that the resulting constraint
    $22y + 8z \cleq 15$ can be normalised by dividing by
    $gcd(22,8)=2$, resulting into $11y + 4z \cleq 15/2$ and, by
    rounding down, $11y + 4z \cleq 7$.\comment{\qed}}
 \end{example}

\section{\warning{\IntSat}}
\label{from-sat-to-ilp}

{It is well known that the problem of SAT can be viewed as a
  particular case of ILP in which there is no objective function, all
  variables are binary and each constraint is of the form
  $x_1 +\ldots +x_m -y_1 \ldots -y_n > -n$, which is an equivalent
  formulation of a clause $x_1 \mios x_m \mio \no{y_1} \mios \no{y_n}$
  using that $\no{y_j} = 1 - y_{j}$. This observation has been used in
  previous work as a starting point for the generalisation of CDCL to
  ILP \cite{JovanovicDeMouraJAR2013}. However, a major obstacle that
  researchers have encountered in this extension is the so-called
  \emph{rounding problem}, to which we devote Section
  \ref{rounding-problem}.}

\subsection{The Rounding Problem}
\label{rounding-problem}

{In order to describe the rounding problem, first of all let
  us review how it has been typically attempted to generalise CDCL
  from SAT to ILP.}

{In ILP, variables are no longer restricted to be just binary
  and may be general integer. As a consequence, the notion of literal
  of SAT needs to be adapted. In the ILP context, \emph{bounds}
  may
  play the role of literals. Namely, lower bounds of the form $a \bleq x$
  correspond to positive literals, whereas upper bounds $x \bleq a$
  correspond to negative literals. Note that, if $x$ is a binary
  variable, and therefore the bounds $0 \bleq x$ and $x \bleq 1$ must
  always hold, then $1 \bleq x$ forces $x$ to take value $1$ (true), while
  $x \bleq 0$ forces $x$ to take value $0$ (false).}

{Moreover, in ILP constraints can be arbitrary integer linear
  constraints instead of purely clauses. Therefore the propagation
  mechanism has to be extended. In this setting, (Boolean) propagation
  can be replaced with \emph{bound propagation}. For the sake of
  simplicity let us define it by example, and leave a formal
  presentation for Section \ref{bound-propagation}.}

\begin{example}
  { Let us see an example of bound propagation. By
    transitivity, from the lower bound $1 \bleq x$, the upper bound
    $y \bleq 2$, and the constraint $x - 2 y + 5 z \cleq 5$, we infer
    that $1-4+5z \cleq 5$, as $y \bleq 2$ implies $-4 \bleq -2y$.
    Simplifying we obtain $5z \cleq 8$, and hence $z\cleq 8/5$, and by
    rounding down, $z \bleq 1$.}
%
\comment{\qed}\end{example}

{Now that it has been outlined how to propagate bounds, in
  order to generalise the CDCL algorithm it remains to be seen how to
  trace back the propagations when a conflict occurs. To that end, one
  can see the propositional resolution rule used in SAT as a means to
  eliminate a variable from a conjunction of two clauses. When
  considering integer linear constraints, a natural candidate for
  playing the same role is the \emph{cut rule}. However, as the
  following example illustrates, mimicking the conflict analysis of
  CDCL SAT solvers does not yield the expected result.}





\begin{example}
\label{exrounding}
{Assume we have two constraints $x + y + 2 z \cleq 2$ and
  $x + y - 2 z \cleq 0$. Let us first take the decision $0 \bleq x$,
  which propagates nothing. Then we take another decision $1 \bleq y$,
  which due to $x + y + 2 z \cleq 2$ propagates $z \bleq 0$ (since
  $0+1+2z \cleq 2$, we get $2z\cleq 1$, that is $z \bleq 1/2$, and by
  rounding down finally $z \bleq 0$). Then $x +y - 2 z \cleq 0$
  becomes a \emph{conflict}: it is false in the current
  assignment $\ A = (\ 0 \bleq x, \ 1 \bleq y, \ z \bleq 0 \ )$, as
  $0 \bleq x, \ 1 \bleq y$ and $z \bleq 0$ imply that
  $1 \bleq x +y - 2 z$.}

{Now let us attempt a straightforward generalisation of
  conflict analysis: as $z \bleq 0$ is the topmost (last propagated)
  bound, we apply a cut inference eliminating $z$ between
  $x + y + 2 z \cleq 2$, the \emph{reason constraint} of the
  propagation, and $x + y - 2 z \cleq 0$, which is now a
  \emph{conflicting constraint} playing the analogous role of a
  conflicting clause. By adding these two constraints we obtain a new
  constraint $2 x + 2y \cleq 2$, or equivalently, $x + y \cleq 1$.
  Then the conflict analysis is over because there is only one bound
  in $A$ that is relevant for the conflicting constraint and which is
  at, or above, the last decision, namely $1 \bleq y$. But
  unfortunately at this point the conflicting constraint, that is
  $x + y \cleq 1$, is no longer false in $A$, breaking (what should
  be) the invariant. Hence, from $0 \bleq x$ it only propagates
  $y \bleq 1$, which is weaker than $y \bleq 0$, the negation of the
  previous decision $1 \bleq y$, which was expected to reverse now. As
  a consequence, the constraint that should be learnt is \emph{too
    weak to justify a backjump}. This problem is due to the rounding
  that takes place when propagating $z$ \footnote{See Lemma
    \ref{no-rounding} in Section \ref{bound-propagation} for a formal
    justification.}). \comment{\qed}}
\end{example}

{The rounding problem illustrated in Example~\ref{exrounding}
  was addressed in an ingenious way by Jovanovi\'c and de Moura in
  their \CutSat procedure \cite{JovanovicDeMouraJAR2013}. In
  this work, a decision can only make a variable \emph{equal} to its
  current lower or upper bound; i.e., if $x$ is the decision variable
  and its current domain is determined by the lower bound $l \bleq x$
  and the upper bound $x \bleq u$, then the next decision has to be
  either $x \bleq l$ (thus fixing the value of $x$ to $l$) or
  $u \bleq x$ (fixing the value of $x$ to $u$). Although, on the one
  hand, this restrains the decision heuristics significantly, on the
  other it is necessary so as to compute, at each conflict caused by
  bound propagations with rounding, \emph{tightly propagating}
  constraints that also explain the same propagations but
  \emph{without} rounding. Conflict analysis can then be performed
  using these tightly propagating constraints only, and thanks to that, the
  resulting constraint is guaranteed to justify a backjump, as in the
  SAT case. Nonetheless, there is yet another toll to be paid: the
  termination condition of the conflict analysis in
  \cite{JovanovicDeMouraJAR2013} requires eliminating the variables of
  the propagated bounds until a decision bound allows one to stop. In
  SAT, the analogous condition would require that, unlike in the 1UIP
  learning scheme, the literal $l_{top}$ in the CDCL algorithm should
  be a decision. The resulting learning scheme is known as the AllUIP
  scheme \cite{Zhangetal2001ICCAD}, and is well-known to perform
  very poorly, 
  significantly worse than the 1UIP one.}

\medskip

\warning{ In what follows, we will present \IntSat, a new family of
  algorithms that extend the conflict-driven
  \emph{clause}-learning scheme from SAT to a conflict-driven
  \emph{constraint}-learning scheme\footnote{{We will be using the same
    acronym CDCL for both when there is no ambiguity.}} in ILP with an alternative
  solution to the rounding problem. However,
  unlike \cite{JovanovicDeMouraJAR2013}, these algorithms admit
  arbitrary new bounds as decisions, and guide the search exactly as
  with the 1UIP approach in SAT solving.}

\subsection{Bound Propagation}
\label{bound-propagation}

{Here we formally define bound propagation,
  which is a key subprocedure in the \warning{\IntSat} CDCL algorithms that will be
  introduced later on.}

Let $A$ be a set of bounds.
We call two bounds $a\bleq x$ and $x \bleq a'$ \emph{contradictory} if $a>a'$.
A bound $a \bleq x$ is \emph{redundant} with another bound
$a' \bleq x$ if $a' \geq a$. Similarly, $x \bleq a$ is redundant with
another bound $x \bleq a'$ if $a' \leq a$. If bound $B$ is redundant
with another bound $B'$, then we say $B'$ is \emph{stronger} than $B$.
By $\min_A(a_1 x_1 +\ldots+ a_nx_n)$ and
$\max_A(a_1 x_1 +\ldots+ a_nx_n)$ we represent respectively the minimum
and the maximum values of the expression $a_1 x_1 +\ldots+ a_nx_n$
subject to the condition that $x_1, \ldots, x_n$ satisfy the bounds in $A$.

\begin{definition}[False constraint, conflict]
Let $A$ be a set of bounds.
If $C$ is a constraint such that $\{C\} \cup A$ has no solution, then
we say that $C$ is \emph{false} or a \emph{conflict} in $A$.
\end{definition}

The following lemma provides us with a characterisation of conflicts:

\begin{lemma}
  \label{lemma:conflicts}
  Let $C$ be a constraint of the form $a_1 x_1 + \ldots + a_nx_n \cleq a_0$ and
  $A$ be a set of pairwise non-contradictory bounds. The following hold:

  \begin{enumerate}

  \item Let $lb_{i}$ be the strongest lower bound of $x_{i}$ in $A$
    (or $-\infty$ if there are none), and let $ub_{i}$ be the strongest
    upper bound of $x_{i}$ in $A$ (or $+\infty$ if there are none).
    Then

    $$
    \min_{A}(a_{i} x_{i}) =
    \left\{
    \begin{array}{lll}
      a_{i} \cdot lb_{i} & \quad & \hbox{ if } a_{i} > 0\ ,\\
      a_{i} \cdot ub_{i} & & \hbox{ if } a_{i} < 0\ .\\
    \end{array}
    \right.
    $$

\medskip

\item $C$ is false in $A$ if and only if $$\sum_{i} \min_{A}(a_{i} x_{i}) > a_{0}\ .$$

  \end{enumerate}

\end{lemma}

\begin{proof}
  See Appendix \ref{app:bound-propagation}.
\end{proof}

\medskip

{If $C$ is a constraint and $R$ is a set of bounds, then $C$ and $R$
can be used to propagate new bounds as the following lemma indicates:}

\begin{lemma}
  \label{lemma:propagations}
  Let $C$ be a constraint $a_1 x_1 + \ldots + a_nx_n \cleq a_0$ and
  $R$ a set of pairwise non-contradictory bounds. Let $x_{j}$ be a
  variable and define
  $e_{j} = (a_{0} - \sum_{i \neq j} \min_{R}(a_{i} x_{i})) / a_{j}$. The following hold:

  \begin{enumerate}
  \item If\; $a_{j} > 0$\; then\; $\{ C \} \cup R \;\models\; x_{j} \bleq \lfloor e_{j} \rfloor$.
  \item If\; $a_{j} < 0$\; then\; $\{ C \} \cup R \;\models\; \lceil e_{j} \rceil \bleq x_{j}$.
  \end{enumerate}

\end{lemma}

\begin{proof}
  See Appendix \ref{app:bound-propagation}.
\end{proof}

The previous lemma motivates the following definition, which presents
the core concept in this subsection:

\begin{definition}[Bound propagation]
  {Let $C$, $R$, $x_{j}$ and $e_{j}$ be as in Lemma
    \ref{lemma:propagations}. We say that $C$ and $R$ \emph{propagate}
    bound $x_{j} \bleq \lfloor e_{j} \rfloor$ if $a_{j} > 0$, or bound
    $\lceil e_{j} \rceil \bleq x_{j}$ if $a_{j} < 0$.}
\end{definition}

\warning{Finally, we conclude this subsection with a lemma that implies that
the problem exposed in Section \ref{rounding-problem} -- that is, that
the constraint to be learnt may be too weak to explain a backjump, is
indeed due to rounding (hence the name \emph{rounding problem}).
Namely, we will prove that if a constraint $C_{2}$ propagates a bound
and constraint $C_{1}$ is false with that bound, and the cut rule is
applied between the two without rounding, then the new constraint
$C_{3}$ that is obtained is still false. As a consequence, the falsity
of the conflicting constraint in conflict analysis is kept invariant,
and therefore in the absence of rounding in bound propagation the
rounding problem cannot arise.}

\begin{lemma}
  \label{no-rounding}
  {
  Let $R$ be a set of non-contradictory bounds, $C_{1}$ be a
  constraint of the form $a_1 x_1 + \ldots + a_nx_n \cleq a_0$, and
  $C_{2}$ a constraint of the form
  $b_1 x_1 + \ldots + b_nx_n \cleq b_0$. Let us assume $a_{j} < 0$,
  $b_{j} > 0$ and that $C_{1}$ is false in
  $R \cup \{ x_{j} \bleq e_{j}\}$, where
  $e_{j} = (b_{0} - \sum_{i \neq j} \min_{R}(b_{i} x_{i})) / b_{j} \in
  \ints$. Let us also assume that $x_{j} \bleq e_{j}$ is the strongest
  upper bound of $x_{j}$ in $R \cup \{ x_{j} \bleq e_{j}\}$. Let
  $C_{3}$ be the result of applying a cut inference between $C_{1}$
  and $C_{2}$ eliminating $x_{j}$:
  $$ C_{3}: (a_{1} b_{j} - a_{j} b_{1}) x_{1} + \ldots + (a_{n} b_{j} - a_{j} b_{n}) x_{n} \cleq a_{0} b_{j} - a_{j} b_{0}\ .$$
  Then $C_{3}$ is false in $R$.
  }
\end{lemma}

\begin{proof}
  See Appendix \ref{app:bound-propagation}.
\end{proof}

{The symmetric version of the lemma when $a_{j} > 0, b_{j} < 0$ and
  the propagated bound is a lower bound can be stated and proved
  analogously.}

\subsection{\warning{The Core \IntSat Algorithm}}
\label{intsat}

{As outlined in Section \ref{intro}, in this work we introduce two new
  CDCL algorithms for ILP. Although they are different in the way
  conflict analysis, backjumping and learning are performed, they
  share the same structure. \warning{This common part,
    which will henceforth be referred to as \warning{\IntSat},
    is presented in this
  section.} In the description below, this \emph{core} algorithm just
  decides the existence of integer solutions, i.e., it considers
  feasibility problems only. Optimisation, as well as other
  extensions, will be discussed later on in Section \ref{extensions}.}

{First of all, let us introduce the following definitions. Let $A$ be
  a sequence of bounds over a set of variables $X$. A variable $x$ is
  \emph{defined to $a$} in $A$ if $a\bleq x \in A$ and
  $x \bleq a \in A$ for some $a$. Note that if all variables of $X$
  are defined and there are no contradictory bounds in $A$ then $A$
  can be seen as a total assignment $A\colon X \to \ints$. A bound $B$
  is \emph{fresh} in $A$ if there is no bound $B' \in A$ such that $B$
  is redundant or contradictory with $B'$.}

{The main idea of the core \warning{\IntSat} algorithm is as follows. The assignment
  $A$, implemented as a stack of bounds, is initially empty. Bound
  propagation is applied exhaustively while no conflict is detected.
  Every time a constraint $RC$ and a set of bounds $RS$ with
  $RS\subseteq A$ propagate a new bound $B$, this bound is pushed onto
  $A$, associating $RC$ and $RS$ to $B$ as the \emph{reason
    constraint} and the \emph{reason set} of $B$, respectively. As in
  the SAT case, when no more propagations are possible and there is no
  conflict, if all variables are defined then the assignment
  determines a solution and the algorithm terminates, otherwise a
  decision is made. On the other hand, when a conflict arises, if
  there are no decisions to be undone we can conclude that there is no
  solution, otherwise a subprocedure performs conflict analysis,
  backjumping and learning. As a result, new constraints may be added
  to the set of constraints, and the assignment undoes the last
  decision (and possibly more) and is extended with a new bound. Then
  the process is repeated with a new round of propagations.}

More succinctly, the following pseudo-code describes the core \warning{\IntSat} algorithm
for finding a solution to a set of constraints $S$:

\begin{enumerate}
\item
\textsf{Propagate}: while possible and no conflict appears,
if
$RC$ and $RS$ propagate some fresh bound $B$,
for some constraint $RC$ and set of bounds $RS$ with $RS\subseteq A$,
then push $B$ onto $A$,
associating to $B$ the \emph{reason constraint} $RC$ and the
\emph{reason set} $RS$.
\item
\textbf{if} there is no conflict\\
  \snp \textbf{if} all variables are defined in $A$, output `solution $A$' and halt.\\
  \snp \textbf{else} \textsf{Decide}: push a fresh bound $B$, marked
  as a \emph{decision}, and go to step 1.

\item \textbf{if} there is a conflict and $A$ contains no decisions,\\
  \snp output `infeasible' and halt.

\item \textbf{if} there is a conflict and $A$ contains some decision,\\
  \snp \textsf{Conflict\_analysis\_Backjump\_Learn}: compute a new assignment $A'$ and a set $T$
  \\ \snp of new constraints to be added to the set of constraints $S$.
  \\ \snp
  Replace $A$ by
  $A'$, add the constraints in $T$ to $S$, and go to step 1.

\end{enumerate}

\noindent
\warning{Unlike in our exposition of CDCL for SAT in Section \ref{sat}, in the
above algorithm conflict analysis, backjumping and learning have been
unified as a single subprocedure
\textsf{Conflict\_analysis\_Backjump\_Learn}. This subroutine is not developed
explicitly until Sections \ref{set-based} and \ref{cut-based}, which
will present
two different ways in which it can be concreted,
although many possibilities exist.} The only requirements that should
be satisfied are summarised in the following definition:

\begin{definition}
  Procedure \textsf{Conflict\_analysis\_Backjump\_Learn} with input a
  set of constraints $S$ and an assignment $A$ such that $S$ contains
  a conflict with $A$ and $A$ contains a decision, and output a set
  of constraints $T$ and an assignment $A'$, is \emph{valid} if:

  \begin{enumerate}

  \item
  \label{def:valid:termination}
    It terminates: the output is computed in finite time.

  \item
    \label{def:valid:consequence}
Any constraint $C \in T$ is a consequence of $S$, that is,
    $S \models C$.

  \item
  \label{def:valid:decomposition}
  The assignment $A$ can be decomposed as $N\, D\, M$, where $N$
    and $M$ are sequences of bounds and $D$ is a decision bound, and
    $A'$ is of the form $N\, B$, where $B$ is a fresh bound in $N$.

  \item
    \label{def:valid:reason-set}
 Bound $B$ has a \emph{reason set} $RS \subseteq N$ such that
    $S \cup RS \models B$.

  \item
\label{def:valid:reason-constraint}
If bound $B$ has a \emph{reason constraint} $RC$ then $S \models RC$.

  \end{enumerate}

\end{definition}

Non-decision bounds obtained with bound propagation have both a reason
set and a reason constraint. On the contrary, a non-decision bound $B$
produced by the procedure \textsf{Conflict\_analysis\_Backjump\_Learn}
always has a reason set $RS$, but may have a reason constraint $RC$ or
not. Moreover, even if it is defined, this reason constraint is
\emph{only} required to be a consequence of $S$. Although for
practical considerations it may be convenient that $RC$ propagates
$B$, the correctness of the algorithm does not depend on this. Finally
also note that, on the other hand, decision bounds never have either a
reason set or a reason constraint.

\begin{example}
\label{example:core}
Consider the following initial constraints:
$$
\begin{array}{rrrrrrrrrr}
C_0: &       - x  -  y  -  z &\cleq& -2 \\ %
C_1: &         x  +  y       &\cleq&  1 \\
C_2: &         x        +  z &\cleq&  1 \\
C_3: &               y  +  z &\cleq&  1 \\
C_4: &                     x &\cleq&  1 \\
C_5: &                     y &\cleq&  1 \\
C_6: &                     z &\cleq&  3 \\
\end{array}
$$


By adding the initial bounds to the empty stack and propagating
exhaustively, the stack depicted below is obtained:

%
$$
\begin{array}{|c|c|c|}
 & &  \\
       0 \bleq  z    & \;\;\; \{\; x \bleq 1,\; y \bleq 1\;\} \;\;\; & \;C_0\; \\ \hline
      -2 \bleq  y    & \;\;\; \{\; x \bleq 1,\; z \bleq 3\;\} \;\;\; & \;C_0\; \\ \hline
  \;\;-2 \bleq x\;\; & \;\;\; \{\; y \bleq 1,\; z \bleq 3\;\} \;\;\; & \;C_0\; \\ \hline


       x \bleq  1    & \;\;\; \emptyset \;\;\; & \;C_4\; \\ \hline
       y \bleq  1    & \;\;\; \emptyset \;\;\; & \;C_5\; \\ \hline
       z \bleq  3    & \;\;\; \emptyset \;\;\; & \;C_6\; \\ \hline

\multicolumn{1}{c}{\;\;\mathit{bound}\;\;}&
\multicolumn{1}{c}{\;\;\mathit{reason\ set}\;\;}&
\multicolumn{1}{c}{\;\;\mathit{reason\ constraint}\;\;}
\end{array}
$$

Now no more new propagations can be made and there is no
conflict. Since for example variable
$x$ is not defined yet, let us decide $1 \bleq x$. This propagates $y
\bleq 0$ due to $C_{1}$, and $z \bleq 0$ due to
$C_{2}$, leading to the following stack:

$$
\begin{array}{|c|c|c|}
  & &  \\
  \;\;z \bleq 0\;\; & \;\;\; \{\; 1 \bleq x\;\} \;\;\; & \;C_2\; \\ \hline
  \;\;y \bleq 0\;\; & \;\;\; \{\; 1 \bleq x\;\} \;\;\; & \;C_1\; \\ \hline
1 \bleq  x    & \multicolumn{2}{c|}{decision}\\ \hline

       0 \bleq  z    & \;\;\; \{\; x \bleq 1,\; y \bleq 1\;\} \;\;\; & \;C_0\; \\ \hline
      -2 \bleq  y    & \;\;\; \{\; x \bleq 1,\; z \bleq 3\;\} \;\;\; & \;C_0\; \\ \hline
  \;\;-2 \bleq x\;\; & \;\;\; \{\; y \bleq 1,\; z \bleq 3\;\} \;\;\; & \;C_0\; \\ \hline

       x \bleq  1    & \;\;\; \emptyset \;\;\; & \;C_4\; \\ \hline
       y \bleq  1    & \;\;\; \emptyset \;\;\; & \;C_5\; \\ \hline
       z \bleq  3    & \;\;\; \emptyset \;\;\; & \;C_6\; \\ \hline

\multicolumn{1}{c}{\;\;\mathit{bound}\;\;}&
\multicolumn{1}{c}{\;\;\mathit{reason\ set}\;\;}&
\multicolumn{1}{c}{\;\;\mathit{reason\ constraint}\;\;}
\end{array}
$$

Notice that $C_0$ is a conflict, as $C_{0} \cup \{\; x \bleq 1, \; y
\bleq 0, \; z \bleq 0
\;\}$ is infeasible. Since the stack contains the decision $1 \bleq
x$, at this point procedure
\textsf{Conflict\_analysis\_Backjump\_Learn} would be called. In
Sections \ref{set-based} and \ref{cut-based} we will resume the
execution of this example with different implementations of this
procedure. \comment{\qed}
\end{example}

The following theorem states the termination, soundness and
completeness of the core \warning{\IntSat} algorithm, provided the subprocedure that
performs conflict analysis, backjumping and learning is valid:

\begin{theorem}
  \label{theorem:termination-soundness-completeness}
  Let us assume procedure \textsf{Conflict\_analysis\_Backjump\_Learn}
  is valid. Then the core \warning{\IntSat} algorithm, when given as input a finite set
  of constraints $S$ including for each variable $x_i$ a lower bound
  $lb_i \bleq x_i$ and an upper bound $x_i \bleq ub_i$, always
  terminates, finding a solution if, and only if, there exists one,
  and returning `infeasible' if, and only if, $S$ is infeasible.
\end{theorem}

\begin{proof}
  See Appendix \ref{app:theorem:termination-soundness-completeness}.
\end{proof}

\subsection{Resolution-based \IntSat}
\label{set-based}

In this section we develop a possible way of concreting the
procedure
\textsf{Conflict\_}\textsf{analysis\_}\textsf{Backjump\_Learn}
introduced in Section \ref{intsat} as a subroutine of the core
\warning{\IntSat} algorithm. The main idea is to mimic the resolution-based conflict
analysis in SAT by only using \emph{reason sets}, which are viewed now
as negations of clauses of bounds.
Thus, for now no reason constraints are considered in this purely resolution-based conflict analysis (until the hybrid versions
explained in the next subsection).
%
%
One of the shortcomings of
this approach is that, at the end of conflict analysis, the
explanation of the backjump is not a constraint but a clause of
bounds, which does not belong to the language of ILP and hence cannot
be learned directly. For this reason here we also review techniques
that, in some common situations, allow one to convert clauses of
bounds into equivalent constraints. If successful, a new constraint
justifying the backjump is learned.
\warning{Otherwise, in the cases in which this equivalent constraint
  cannot be found, nothing is learned (or at best only a \emph{weaker}
  constraint could be learned). In any case, this does not affect the
  validity of the procedure and backjumping can be performed anyway.}


A more precise pseudo-code description of
\textsf{Conflict\_}\textsf{analysis\_}\textsf{Backjump\_Learn}
following this idea is shown below. The input consists of a set of
constraints $S$ and an assignment $A$ such that $S$ contains a
conflict with $A$ and $A$ contains a decision, and the output is a set
$T$ of constraints that can be learned and a new assignment $A'$.

\begin{enumerate}[label=\arabic*.]
\item
  \textsf{Conflict analysis}:
  Let us define $CS$ as the \emph{Conflicting Set} of bounds.
  Initially, $CS$ is a subset of bounds of $A$ causing the falsehood
  of $C$ for a certain conflict $C$ in $S$, i.e, $C \in S$ is false in
  $CS \subseteq A$.

  \underline{Invariants:} $CS \subseteq A$ and $S\cup CS$ is infeasible.

  \smallskip

  Let $B$ be the bound in $CS$ that is topmost in $A$, and $RS$ its
  reason set. Replace $CS$ by
  $(CS\setminus \{B\}) \cup RS$. Repeat this until $CS$ contains a
  single bound $B_{top}$ that is, or is above, $A$'s topmost decision.

\item
\textsf{Backjump}: Assign $A'$ a copy of $A$. Pop bounds from $A'$ until
either there are no decisions in $A'$ or,
for some $B$ in $CS$ with $B\not=B_{top}$, there are no decisions
above $B$ in $A'$.
Then push $\no{B_{top}}$ onto $A'$ with associated reason set $CS \setminus \{B_{top}\}$.

\item \textsf{Learn}: if $\no{CS} = \{ \no B \;|\; B \in CS \}$, viewed as a clause of bounds, can be
  converted into \warning{an equivalent} constraint $CC$, let $T=\{CC\}$
else  $T = \emptyset$.

\end{enumerate}

Next let us argue that, as claimed in the introduction to this
section, procedure
\textsf{Conflict\_}\textsf{analysis\_}\textsf{Backjump\_Learn} is
reproducing the conflict analysis from SAT based on resolution. To
that end, it is convenient to view the algorithm as working on the
\emph{logical negations} of the conflicting set and the reason sets.
Namely, one of the invariants of conflict analysis is that $S\cup CS$
is infeasible. Equivalently, in a more logical notation we can say
that $S \land (\bigwedge_{B \in CS} B)$ is unsatisfiable, or that
$S \models \bigvee_{B \in CS} \no{B}$. Similarly, the reason set of a
non-decision bound $B'$ satisfies that $S \cup RS \models B'$, which
can be reformulated logically as
$S \models B' \lor \bigvee_{B \in RS} \no{B}$. Drawing a parallel with
SAT, the clause $\bigvee_{B \in CS} \no{B}$ would play the role of the
\emph{conflicting clause}, while the clause $\bigvee_{B \in RS} \no{B}$
would correspond to the \emph{reason clause} of $B'$. From this
viewpoint, let us consider the step in the above algorithm in which
the propagation of bound $B'$ with reason set $RS$ is unfold and $CS$
is replaced by $(CS\setminus \{B'\}) \cup RS$. Now this can be
interpreted as a resolution inference eliminating `literal' $B'$
between clause $\bigvee_{B \in CS} \no{B}$, viewed as
$\no{B'} \lor (\bigvee_{B \in CS \setminus \{B'\}} \no{B})$, and
clause $B' \lor (\bigvee_{B \in RS} \no{B})$. Note that the resolvent
$$(\bigvee_{B \in CS \setminus \{B'\}} \no{B}) \lor (\bigvee_{B \in RS}
\no{B})$$ is precisely the negation of (the conjunction of bounds in)
the set $(CS\setminus \{B'\}) \cup RS$.


\begin{example}
  Let us revisit Example \ref{exrounding}. There are two constraints
  $x + y + 2 z \cleq 2$ and $x + y - 2 z \cleq 0$. We take the
  decision $0 \bleq x$, which propagates nothing, and later on another
  decision $1 \bleq y$, which due to $x + y + 2 z \cleq 2$ propagates
  $z \bleq 0$. This bound is then pushed with associated reason set
  $\{\; 0 \bleq x, \ 1 \bleq y \;\}$, resulting into the stack
  $\ A = (\ 0 \bleq x, \ 1 \bleq y, \ z \bleq 0 \ )$. Now
  $x +y - 2 z \cleq 0$ is a conflict, and
  $CS = \{\; 0 \bleq x,\ 1 \bleq y,\ z \bleq 0\;\} \subseteq A$ is a
  set of bounds causing its falsehood.

  Let us apply the procedure
  \textsf{Conflict\_}\textsf{analysis\_}\textsf{Backjump\_Learn}. In
  the first iteration of conflict analysis we unfold the propagation
  of $z \bleq 0$, and then $CS$ becomes
  $CS \setminus \{ z \bleq 0 \} \cup \{ 0 \bleq x, 1 \bleq y\} = \{ 0
  \bleq x, 1 \bleq y\}$. After this step there is a single bound of
  $CS$ which is, or is above, the assignment's topmost decision
  $1 \bleq y$ (namely $1 \bleq y$ itself, thus playing the role of
  $B_{top}$), and therefore \textsf{Conflict analysis} concludes. Then
  \textsf{Backjump} starts, and pops $z \bleq 0$ and $1 \bleq y$ from
  the assignment. Since $0 \bleq x$ is a bound in $CS$ different from
  $1 \bleq y$ such that there are no decisions above it in the current
  assignment $(\ 0 \bleq x\ )$, no more bounds are popped. Finally
  $y \bleq 0$ (that is, $\no{B_{top}}$) is pushed
  onto the stack with reason set
  $\{ 0 \bleq x, 1 \bleq y\} \setminus \{1 \bleq y \} = \{ 0 \bleq x
  \}$, resulting into the assignment $(\ 0 \bleq x,\ y \bleq 0\ )$.

  In this case the clause of the negations of the bounds in the final
  conflicting set, namely
  $$\no{0 \bleq x} \;\lor\; \no{1 \bleq y} \;\;\;\equiv\;\;\; x \bleq -1 \;\lor\; y \bleq 0\ ,$$
  cannot be converted into a constraint, as can be seen e.g. by
  \warning{geometric} arguments, and therefore nothing is learnt (i.e.,
  $T = \emptyset$). However, if additionally variable $y$ were binary
  and $x$ had an upper bound $x \bleq u$ for a certain $u \in \ints$
  in the initial set of constraints, \warning{then one can see that
  the clause $ x \bleq -1 \,\lor\, y \bleq 0$ could equivalently be transformed into
  the constraint $x \cleq u -uy -y$ (and finally
  $T = \{ x \cleq u -uy -y \}$); see the end of this section for a systematic way of obtaining these transformations.} \comment{\qed}
\end{example}

\begin{example}
\label{example:set}
  Let us resume the execution of the core \warning{\IntSat} algorithm from Example
  \ref{example:core}. The constraints are:
$$
\begin{array}{rrrrrrrrrr}
C_0: &       - x  -  y  -  z &\cleq& -2 \\ %
C_1: &         x  +  y       &\cleq&  1 \\
C_2: &         x        +  z &\cleq&  1 \\
C_3: &               y  +  z &\cleq&  1 \\
C_4: &                     x &\cleq&  1 \\
C_5: &                     y &\cleq&  1 \\
C_6: &                     z &\cleq&  3 \\
\end{array}
$$

Let $A_{0}$ be the stack corresponding to decision level $0$:

$$
\begin{array}{|c|c|c|}
 & &  \\
       0 \bleq  z    & \;\;\; \{\; x \bleq 1,\; y \bleq 1\;\} \;\;\; & \;C_0\; \\ \hline
      -2 \bleq  y    & \;\;\; \{\; x \bleq 1,\; z \bleq 3\;\} \;\;\; & \;C_0\; \\ \hline
  \;\;-2 \bleq x\;\; & \;\;\; \{\; y \bleq 1,\; z \bleq 3\;\} \;\;\; & \;C_0\; \\ \hline


       x \bleq  1    & \;\;\; \emptyset \;\;\; & \;C_4\; \\ \hline
       y \bleq  1    & \;\;\; \emptyset \;\;\; & \;C_5\; \\ \hline
       z \bleq  3    & \;\;\; \emptyset \;\;\; & \;C_6\; \\ \hline

\multicolumn{1}{c}{\;\;\mathit{bound}\;\;}&
\multicolumn{1}{c}{\;\;\mathit{reason\ set}\;\;}&
\multicolumn{1}{c}{\;\;\mathit{reason\ constraint}\;\;}
\end{array}
$$

In Example \ref{example:core} the execution was suspended at the first
conflict that was encountered. Namely, the stack is

$$
\begin{array}{|c|c|c|}
  & &  \\
  \;\;z \bleq 0\;\; & \;\;\; \{\; 1 \bleq x\;\} \;\;\; & \;C_2\; \\ \hline
  \;\;y \bleq 0\;\; & \;\;\; \{\; 1 \bleq x\;\} \;\;\; & \;C_1\; \\ \hline
1 \bleq  x    & \multicolumn{2}{c|}{decision}\\ \hline

       \multicolumn{3}{|c|}{A_{0}}\\ \hline


\multicolumn{1}{c}{\;\;\mathit{bound}\;\;}&
\multicolumn{1}{c}{\;\;\mathit{reason\ set}\;\;}&
\multicolumn{1}{c}{\;\;\mathit{reason\ constraint}\;\;}
\end{array}
$$

and the conflict is $C_0$, which is false due to $\{\; x \bleq 1, \; y
\bleq 0, \; z \bleq 0
\;\}$.

Let us apply \textsf{Conflict\_}\textsf{analysis\_}\textsf{Backjump\_Learn}.
We start conflict analysis with the conflicting set $CS = \{\; x \bleq 1, \;
y \bleq 0, \; z \bleq 0 \;\}$. The bound in
$CS$ which is topmost in the stack is $z \bleq
0$, which has reason set $\{\; 1 \bleq x\; \}$. Therefore the new
$CS$ becomes $\{\; x \bleq 1, \; y \bleq 0, \; z \bleq 0 \;\}
\setminus \{\; z \bleq 0 \;\} \cup \{\; 1 \bleq x \;\} = \{\; x \bleq 1, \; y
\bleq 0, \; 1 \bleq x
\;\}$. Notice that there are still two bounds in
$CS$ of the last decision level, namely $y \bleq 0$ and $1 \bleq
x$. Of these, the topmost one is $y \bleq 0$, which has reason set $\{\; 1 \bleq x \;\}$. Then the new conflicting set is $\{\; x \bleq 1, \; y
\bleq 0, \; 1 \bleq x \;\} \setminus \{\; y \bleq 0 \;\} \cup \{\; 1 \bleq x
\;\} = \{\; x \bleq 1, \; 1 \bleq x
\;\}$. Now there is a single bound $B_{top}$, which is $1 \bleq
x$, which is or is above the stack's topmost decision, so conflict
analysis is over.

The next step is \textsf{Backjump}, which pops bounds from the stack
until either there are no decisions or, for some $B$ in $CS$ with
$B\not=B_{top}$, there are no decisions above $B$ in the stack. In
this case the latter holds with bound $x \bleq 1$ playing the role of
$B$ after having popped $z \bleq 0$, $y \bleq 0$ and $1 \bleq x$. Then
$\no{B_{top}}$, that is $x \bleq 0$, is pushed with reason set
$\{\; x \bleq 1, \; 1 \bleq x \;\} \setminus \{ \; 1 \bleq x \;\} =
\{\; x \bleq 1 \;\}$, leading to the stack:

$$
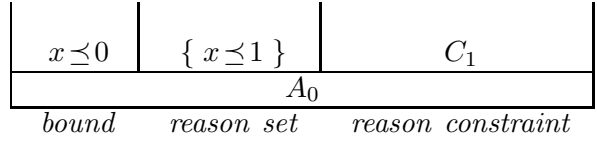
\begin{array}{|c|c|c|}
  & &  \\
  \;\;x \bleq 0\;\; & \;\;\; \{\; x \bleq 1\;\} \;\;\; & \;C_1\; \\ \hline

       \multicolumn{3}{|c|}{A_{0}}\\ \hline


\multicolumn{1}{c}{\;\;\mathit{bound}\;\;}&
\multicolumn{1}{c}{\;\;\mathit{reason\ set}\;\;}&
\multicolumn{1}{c}{\;\;\mathit{reason\ constraint}\;\;}
\end{array}
$$

On the other hand, the clause of the negations of the final conflicting set
$\{\; x \bleq 1, \; 1 \bleq x \;\}$, namely
$$\no{x \bleq 1} \;\lor\; \no{1 \bleq x} \;\;\;\equiv\;\;\; 2 \bleq x \;\lor\; x \bleq 0\ ,$$
cannot be converted into a constraint, and so no constraint can be
learnt.

This concludes the execution of
\textsf{Conflict\_}\textsf{analysis\_}\textsf{Backjump\_Learn}. Back
to the core \warning{\IntSat} algorithm, now due to $C_0$ we have that $x \bleq 0$ and
$y \bleq 1$ propagate $1 \bleq z$ , and $x \bleq 0$ and $z \bleq 3$
propagate $-1 \bleq y$. Then due to $C_{3}$ bound $1 \bleq z$
propagates $y \bleq 0$, and $-1 \bleq y$ propagates $z \bleq 2$. In
turn, thanks to $C_{0}$, we have that $x \bleq 0$ and $y \bleq 0$
propagate $2 \bleq z$, and $x \bleq 0$ and $z \bleq 2$ propagate
$0 \bleq y$. Finally, bounds $2 \bleq z$ and $0 \bleq y$ make $C_{3}$
false. Since there are no decisions left in the stack, the algorithm
terminates reporting that there is no solution. \comment{\qed}
\end{example}

The following theorem ensures that the procedure
\textsf{Conflict\_}\textsf{analysis\_}\textsf{Backjump\_Learn} is
valid, as required by Theorem
\ref{theorem:termination-soundness-completeness}:

\begin{theorem}
  \label{valid-resolution-based}
  The above procedure
  \textsf{Conflict\_}\textsf{analysis\_}\textsf{Backjump\_Learn} is
  valid.
\end{theorem}

\begin{proof}
  See Appendix \ref{app:validity}.
\end{proof}


Finally, there is a step in the above description of the algorithm that
requires further explanation. Namely, let us see when and how the
clause consisting of the negations of the bounds in the final
conflicting set can be converted into an equivalent constraint. For
example, this transformation can be achieved if all bounds involve
binary variables, except for at most one
\cite{DBLP:journals/aicom/AsinBN15}. More specifically, if the clause
is of the form
$$1 \cleq x_{1} \,\lor\, \ldots \,\lor\, 1 \cleq x_{m} \,\lor\, y_{1} \cleq 0 \,\lor\,
\ldots \,\lor\, y_{n} \cleq 0 \,\lor\, k \cleq z\ ,$$ where variables
$x_{1}, \ldots, x_{m}, y_{1}, \ldots, y_{n}$ are binary, $z$ is an
integer variable with lower bound $lb$ (that is, $lb \cleq z$ for any
solution) and $k > lb$ is an integer, then the clause can equivalently be
written as
$$k - (k - lb) \big(\sum_{i=1}^{m}x_{i} + \sum_{j=1}^{n} (1-y_{j})\big) \cleq z\ .$$
%
%
Similarly, if the clause is of the form
$$1 \cleq x_{1} \,\lor\, \ldots \,\lor\, 1 \cleq x_{m} \,\lor\, y_{1} \cleq 0 \,\lor\,
\ldots \,\lor\, y_{n} \cleq 0 \,\lor\, z \cleq k\ ,$$
where variables
$x_{1}, \ldots, x_{m}, y_{1}, \ldots, y_{n}$ are binary, $z$ is an
integer variable with upper bound $ub$ (i.e., $z \cleq ub$ for any
solution) and $k < ub$ is an integer, then the clause can equivalently be
written as
$$z \cleq k + (ub - k) \big(\sum_{i=1}^{m}x_{i} + \sum_{j=1}^{n} (1-y_{j})\big)\ .$$

\subsection{Hybrid Resolution and Cut-based \IntSat}
\label{cut-based}

The algorithm for the procedure
\textsf{Conflict\_}\textsf{analysis\_}\textsf{Backjump\_}\textsf{Learn}
presented in Section \ref{set-based} suffers from some drawbacks. \warning{In
the first place, if the clause with the negations of the bounds in the
conflicting set cannot be converted into an equivalent constraint,
at best only a weaker constraint can be learned}.
This is inconvenient, as learning is widely acknowledged as
one of the key components to the success of SAT solvers. Moreover,
from a proof complexity perspective, the resolution proof system that
underlies the algorithm \cite{PipatsrisawatD09} is known to be less
powerful than, e.g., the cutting planes proof system, based on the cut
rule. For example, for pigeon-hole problems resolution proofs are
exponentially long \cite{Haken1985}, while short polynomial-sized
proofs exist in cutting planes. Unfortunately, pigeon-hole problems
often arise as subproblems in real-life applications, such as
scheduling and timetabling
\cite{DBLP:journals/anor/AchaN14}: how to fit $n$ tasks in less than $n$ time slots, etc.

This section presents an alternative approach for
\textsf{Conflict\_}\textsf{analysis\_}\textsf{Backjump\_Learn}. This
new algorithm incorporates the cut rule, in an attempt to address the
aforementioned issues. However, the cut rule does not replace
resolution but is applied \emph{in parallel}. Hence, in addition to
the conflicting set, a \emph{conflicting constraint} will also be
maintained in conflict analysis, and when unfolding propagations cuts
will be applied between this constraint and the reason constraints,
which unlike in Section \ref{set-based}, will now be taken into
account. In this way we can bypass the rounding problem and ensure
that backjumping will be possible.

Next a description of the new algorithm for
\textsf{Conflict\_}\textsf{analysis\_}\textsf{Backjump\_Learn} is
shown. As in the previous section, the input consists of a set of
constraints $S$ and an assignment $A$ such that $S$ contains a
conflict with $A$ and $A$ contains a decision, and the output is a set
$T$ of constraints that can be learned and a new assignment $A'$.

\begin{enumerate}[label=\arabic*.]
\item
  \textsf{Conflict analysis}:
  Let us define $CC$ as the \emph{Conflicting Constraint}, and
$CS$ as the \emph{Conflicting Set} of bounds.
  Initially, $CC$ is any conflict and $CS$ is a subset of bounds of
  $A$ causing the falsehood of $CC$, i.e, $CC \in S$ is false in
  $CS \subseteq A$.

  \underline{Invariants:} $S \models CC$, and $CS \subseteq A$, and $S\cup CS$ is infeasible.

\smallskip

\textbf{Repeat} the following three steps:
\begin{enumerate}[label=\theenumi\arabic*.]
\item
Let $B$ be the bound in $CS$ that is topmost in $A$, and $RS$ its reason set.
Replace $CS$ by $(CS\setminus \{B\}) \cup RS$.

\item Let $x$ and $RC$ be the variable and the reason constraint of $B$, respectively.



  If $RC$ is defined and there exists a cut eliminating $x$ between
  $CC$ and $RC$ then replace $CC$ by that cut.

\item 
  \textsf{Early Backjump}: If for some maximal $k\in \nats$, after
  popping $k$ bounds from $A$ the last one being a decision, the
  conflicting constraint $CC$ and a subset $RS'$ of bounds of the
  resulting $A$ propagate some fresh bound $B'$, then assign $A'$ a copy
  of $A$, pop $k$ bounds from $A'$, push $B'$ onto $A'$ with
  associated reason constraint $CC$ and reason set $RS'$, and go to
  \textsf{Learn}.
\end{enumerate}
\textbf{until} $CS$ contains a single bound $B_{top}$ that is, or is above, $A$'s topmost decision.

\item
\textsf{Backjump}: Assign $A'$ a copy of $A$. Pop bounds from $A'$ until
either there are no decisions in $A'$ or,
for some $B$ in $CS$ with $B\not=B_{top}$, there are no decisions
above $B$ in $A'$.
Then push $\no{B_{top}}$ onto $A'$ with associated reason constraint $CC$ and reason set $CS \setminus \{B_{top}\}$.

\item \textsf{Learn}: Let $T=\{CC\}$.

\end{enumerate}

\begin{example} Again let us consider Example~\ref{exrounding}, and
  let us apply the procedure
  \textsf{Conflict\_}\textsf{analysis\_}\textsf{Backjump\_Learn} to
  the conflict described there. We recall we have the set of
  constraints $S = \{x + y + 2 z \cleq 2, x + y - 2 z \cleq 0\}$ and
  the assignment $\ A = (\ 0 \bleq x, \ 1 \bleq y, \ z \bleq 0 \ )$,
  where $0 \bleq x$ and $1 \bleq y$ are decisions while
  $z \bleq 0$ is a propagated bound with reason constraint
  $x +y + 2 z \cleq 2$ and reason set $\{ 0 \bleq x, 1 \bleq y\}$. The
  conflict is $x + y - 2 z \cleq 0 \in S$, which is false due to
  $\{\ 0 \bleq x, \ 1 \bleq y, \ z \bleq 0 \ \} \subseteq A$.

  \textsf{Conflict analysis} starts assigning $x +y - 2 z \cleq 0$ to
  $CC$ and the set of bounds
  $\{\; 0 \bleq x,\ 1 \bleq y,\ z \bleq 0\;\}$ to $CS$. In the first
  iteration, in $CS$ we replace $z \bleq 0$ by its reason set
  $\{\; 0 \bleq x, \ 1 \bleq y \;\}$. The resulting $CS$ is
  $\{\; 0 \bleq x, \ 1 \bleq y \;\}$. A cut eliminating $z$ exists
  (see Example~\ref{exrounding}) and $CC$ becomes $x + y \cleq 1$. Then
  conflict analysis is over because $CS$ contains exactly one
  bound $B_{top}$, which is $1\bleq y$, at or above $A$'s topmost
  decision.
  Then \textsf{Backjump} starts, and we pop bounds until for some $B$
  in $CS$ with $B\not=B_{top}$, there are no decisions above $B$ in
  $A$, in this case, until there are no decisions above $0 \bleq x$ in
  $A$. Hence bounds $z \bleq 0$ and $1 \bleq y$ are popped, and after
  that $\no{B_{top}}$, which is $y \bleq 0$, is pushed with reason set
  $\{\; 0 \bleq x \;\}$ and reason constraint $x + y \cleq 1$. Note
  that this reason constraint is not a `good' reason, i.e., it does
  not propagate $y \bleq 0$, but is still a valid consequence of the
  set of constraints.
  Finally, in \textsf{Learn}, the final $CC$, which is $x +y \cleq 1$,
  is assigned to $T$ so that it can be learned. \comment{\qed}
\end{example}

\begin{example}
  Let us revisit Example
  \ref{example:core} and complete the execution of the core \warning{\IntSat} algorithm.
  Let us recall the constraints:
$$
\begin{array}{rrrrrrrrrr}
C_0: &       - x  -  y  -  z &\cleq& -2 \\ %
C_1: &         x  +  y       &\cleq&  1 \\
C_2: &         x        +  z &\cleq&  1 \\
C_3: &               y  +  z &\cleq&  1 \\
C_4: &                     x &\cleq&  1 \\
C_5: &                     y &\cleq&  1 \\
C_6: &                     z &\cleq&  3 \\
\end{array}
$$

Let us define $A_{0}$ as the stack of decision level $0$:

$$
\begin{array}{|c|c|c|}
 & &  \\
       0 \bleq  z    & \;\;\; \{\; x \bleq 1,\; y \bleq 1\;\} \;\;\; & \;C_0\; \\ \hline
      -2 \bleq  y    & \;\;\; \{\; x \bleq 1,\; z \bleq 3\;\} \;\;\; & \;C_0\; \\ \hline
  \;\;-2 \bleq x\;\; & \;\;\; \{\; y \bleq 1,\; z \bleq 3\;\} \;\;\; & \;C_0\; \\ \hline


       x \bleq  1    & \;\;\; \emptyset \;\;\; & \;C_4\; \\ \hline
       y \bleq  1    & \;\;\; \emptyset \;\;\; & \;C_5\; \\ \hline
       z \bleq  3    & \;\;\; \emptyset \;\;\; & \;C_6\; \\ \hline

\multicolumn{1}{c}{\;\;\mathit{bound}\;\;}&
\multicolumn{1}{c}{\;\;\mathit{reason\ set}\;\;}&
\multicolumn{1}{c}{\;\;\mathit{reason\ constraint}\;\;}
\end{array}
$$

The execution in Example \ref{example:core} was interrupted at the first
conflict, with the following stack:

$$
\begin{array}{|c|c|c|}
  & &  \\
  \;\;z \bleq 0\;\; & \;\;\; \{\; 1 \bleq x\;\} \;\;\; & \;C_2\; \\ \hline
  \;\;y \bleq 0\;\; & \;\;\; \{\; 1 \bleq x\;\} \;\;\; & \;C_1\; \\ \hline
1 \bleq  x    & \multicolumn{2}{c|}{decision}\\ \hline

       \multicolumn{3}{|c|}{A_{0}}\\ \hline


\multicolumn{1}{c}{\;\;\mathit{bound}\;\;}&
\multicolumn{1}{c}{\;\;\mathit{reason\ set}\;\;}&
\multicolumn{1}{c}{\;\;\mathit{reason\ constraint}\;\;}
\end{array}
$$

Procedure \textsf{Conflict\_analysis\_Backjump\_Learn} starts with
conflict analysis, where initially the conflicting constraint
$CC$ is $C_0$, and the conflicting set $CS$ is $\{\; x \bleq 1, \; y
\bleq 0, \; z \bleq 0 \;\}$. The bound in
$CS$ which is topmost in the stack is $z \bleq
0$, which has reason set $\{\; 1 \bleq x\;
\}$ and reason constraint
$C_2$. Thus, as in Example \ref{example:set}, the new conflicting set
becomes $\{\; x \bleq 1, \; y \bleq 0, \; 1 \bleq x
\;\}$. Moreover, the cut rule can be applied to $C_2$ and
$C_0$ to eliminate variable $z$, resulting into $C_7: 1 \bleq
y$, which becomes the new conflicting constraint. Now \textsf{Early
  Backjump} can be applied, since after popping $z \bleq 0$, $y \bleq
0$ and $1 \bleq x$ the constraint $1 \bleq
y$ trivially propagates the fresh bound $1 \bleq y$ in $A_0$. So $1
\bleq y$ is pushed with reason constraint $1 \bleq
y$ and reason set $\emptyset$, leading to the stack:

$$
\begin{array}{|c|c|c|}
  & &  \\
  \;\;1 \bleq y\;\; & \;\;\; \emptyset \;\;\; & \;C_7\; \\ \hline

       \multicolumn{3}{|c|}{A_{0}}\\ \hline


\multicolumn{1}{c}{\;\;\mathit{bound}\;\;}&
\multicolumn{1}{c}{\;\;\mathit{reason\ set}\;\;}&
\multicolumn{1}{c}{\;\;\mathit{reason\ constraint}\;\;}
\end{array}
$$
Moreover, \textsf{Learn} adds $1 \bleq y$ to the set of constraints.
This concludes procedure \textsf{Conflict\_analysis\_Backjump\_Learn}.

Now $1 \bleq y$ propagates $x \bleq 0$ due to $C_{1}$, and $z \bleq
0$ due to $C_{3}$. In turn, $x \bleq 0$, $z \bleq 0$ and $y \bleq
1$ make constraint
$C_0$ false. As there are no decisions left in the stack, the
execution of the core \warning{\IntSat} algorithm terminates reporting that the set of
constraints is infeasible. Note that, compared to Example
\ref{example:set}, the possibility to infer and learn $1 \bleq
y$ allows proving infeasibility with many fewer propagations. \comment{\qed}
\end{example}

Finally, the next theorem ensures that procedure
\textsf{Conflict\_}\textsf{analysis\_}\textsf{Backjump\_Learn} is
valid, as required by Theorem
\ref{theorem:termination-soundness-completeness}:

\begin{theorem}
  \label{valid-cut-based}
  The above procedure
  \textsf{Conflict\_}\textsf{analysis\_}\textsf{Backjump\_Learn} is
  valid.
\end{theorem}

\begin{proof}
  See Appendix \ref{app:validity}.
\end{proof}

\subsection{\warning{Extensions of \IntSat}}
\label{extensions}

The \warning{\IntSat} algorithm described in Section \ref{intsat} only decides the
existence of integer solutions. In order to go beyond feasibility
problems, \emph{optimisation} can be handled in the following standard
way. For finding a solution that minimises a linear objective function
$c_1 x_1 +\ldots+ c_nx_n$, each time a new solution $sol$ is found,
the constraint $c_1 x_1 +\ldots+ c_nx_n \cleq c_0$ where $c_0$ is
$c_1 sol(x_1) +\ldots+ c_n sol(x_n)-1$ is added, so as to attempt to
improve the best solution found so far. This triggers a conflict, from
which the search continues. This objective strengthening is repeated until the problem becomes infeasible.
Bound propagations from these successively stronger constraints turn out to
be very effective for pruning. Unlike what happens in propositional
logic, linear constraints are first-class citizens (i.e., belong to
the core language), and therefore adding the constraints generated
from the objective function is straightforward and does not require
further encodings, as it happens in SAT \cite{MiniSAT+}.

Another simplification we have made for the sake of presentation is
that constraints are assumed to be $\cleq$ inequalities with integer
coefficients. However, with trivial transformations it is possible to
tackle more general constraints. Namely, a constraint
$a_1 x_1 +\ldots+ a_nx_n \cgeq a_0$ can be expressed as
$-a_1 x_1 -\ldots- a_nx_n \cleq -a_0$, a constraint
$a_1 x_1 +\ldots+ a_nx_n \ceq a_0$ can be replaced by the two
constraints $a_1 x_1 +\ldots+ a_nx_n \cleq a_0$ and
$a_1 x_1 +\ldots+ a_nx_n \cgeq a_0$, and rational non-integer
coefficients $a/b$ can be removed by multiplying both sides by $b$.

\section{Implementation}
\label{implem}

In this section we describe some aspects of our current implementation
of \warning{\IntSat}. It consists of roughly 10000 lines of
simple C++ code that make heavy use of standard STL data structures.
For instance, a constraint is an STL vector of monomials (pairs of two
\texttt{int}s: the variable number and the coefficient), sorted by
variable number, plus the independent term.
\warning{Coefficients are never larger than $2^{30}$, and cuts producing any
coefficient larger than $2^{30}$ are simply not performed.
In combination with the fact that we use 64-bit integers for intermediate results
during bound propagation, cuts, normalisation, etc.,
this allows us to prevent overflow with a few cheap and simple tests.}

%
%
%

%




STL vectors are also used, e.g., in the implementation of the
assignment. There is a vector, the \emph{bounds vector}, indexed by
variable number, which can return in constant time the current lower
and upper bounds for that variable. It always stores, for each
variable $x_i$, the positions $pl_i$ and $pu_i$ in the stack of its
current (strongest) lower bound and upper bound, respectively. The
stack itself is another STL vector containing at each position three
data fields: a bound, a natural number $pos$, and an \emph{info} field
that includes, among other information, (pointers to) the reason set
and the reason constraint. The value $pos$ is always the position in
the stack of the previous bound of the same type (lower or upper) for
that variable, with $pos=-1$ for initial bounds. When pushing or
popping bounds, these properties are easy to maintain in constant
time. See Figure \ref{fig:example} for an example of bounds vector and
corresponding stack.

\begin{figure}
  \centering
  \begin{tabular}{r|c|c|}
\cline{2-3}
                 &\multicolumn{2}{|c|}{\textbf{Height in stack of}}\\
                 &\multicolumn{2}{|c|}{\textbf{current bound}}\\
\cline{2-3}
&\textbf{ \ Lower \ }&\textbf{Upper}\\
\cline{2-3}
                 \multicolumn{3}{c}{}\\
\cline{2-3}
$x_1$ & 0 & 1 \\
\cline{2-3}
 & $\vdots$ & $\vdots$ \\
\cline{2-3}
$x_7$ & 40 & 31 \\
\cline{2-3}
 & $\vdots$ & $\vdots$ \\
\cline{2-3}
\end{tabular}
\qquad
\qquad
\begin{tabular}{l|c|c|c|}
                 & \multicolumn{3}{|c|}{$\vdots$} \\
\cline{2-4}
$40\;$  & $\;\;5 \bleq x_7\;\;$  & $\;\;23\;\;$       & \emph{ \ info \ } \\ \cline{2-4}
                 & \multicolumn{3}{|c|}{$\vdots$} \\ \cline{2-4}
$31$  & $x_7 \bleq 6$  & $14$       & \emph{info} \\ \cline{2-4}
                 & \multicolumn{3}{|c|}{$\vdots$} \\ \cline{2-4}
$23$  & $2 \bleq x_7$  & $13$       & \emph{info} \\ \cline{2-4}
                 & \multicolumn{3}{|c|}{$\vdots$} \\ \cline{2-4}
$14$  & $x_7 \bleq 9$  & $-1$        & \emph{info} \\ \cline{2-4}
$13$  & $0 \bleq x_7$  & $-1$        & \emph{info} \\ \cline{2-4}
                 & \multicolumn{3}{|c|}{$\vdots$} \\ \cline{2-4}
$1$   & $x_1 \bleq 8$  & $-1$        & \emph{info} \\ \cline{2-4}
$0$   & $0 \bleq x_1$  & $-1$        & \emph{info} \\ \cline{2-4}
\end{tabular}

\caption{Example of bounds vector (left) and stack (right).}
\label{fig:example}
\end{figure}
\bigskip

\subsection{Bound propagation using filters}

Affordably efficient bound propagation is crucial for performance.
In our current implementation,
for each variable $x$ there are two \emph{occurs
  lists}.  The \emph{positive} occurs list for $x$ contains all pairs
$(I_C,a)$ such that $C$ is a linear constraint where $x$ occurs with a
positive coefficient $a$, and the \emph{negative} one contains
the same for occurrences with a negative coefficient $a$.
Here $I_C$ is an index to the \emph{constraint filter} $F_C$ of $C$ in an
array of constraint filters. The filter is maintained cheaply, and one can guarantee that
\emph{$C$ does not propagate anything as long as $F_C \leq 0$}, thus avoiding many
useless (cache-) expensive visits to the actual constraint $C$.
The filtering technique is based on the following lemma:

\begin{lemma}
  \label{lemma:filtering}
  Let $C$ be a constraint of the form
  $a_1 x_1 + \cdots + a_n x_n \cleq a_0$. Let $R$ be the set
  consisting of the current lower bound $lb_i \bleq x_i$ and the
  current upper bound $x_i \bleq ub_i$ of $x_i$, for each variable
  $x_{i}$. Then the following hold:

  \begin{enumerate}

  \item\label{lemma:filtering:3} Constraint $C$
    propagates a non-redundant bound on $x_{j}$ if and only if
    $$-a_{0} + |a_{j}| \,(ub_{j} - lb_{j}) + \sum_{i} \min_{R}(a_{i} x_{i}) > 0\ .$$

  \item\label{lemma:filtering:4} Constraint $C$ propagates (a non-redundant bound) if and only if
    $$-a_{0} + \max_{j}\,(\;|a_{j}| \,(ub_{j} - lb_{j})\;) + \sum_{i} \min_{R}(a_{i} x_{i}) > 0\ .$$

  \end{enumerate}

\end{lemma}

\begin{proof}

\bigskip

For the first claim, first let us assume $a_{j} > 0$. Then by Lemma
\ref{lemma:propagations}, $C$ and $R$
propagate $x_{j} \bleq \lfloor e_{j} \rfloor$, where
$e_{j} = (a_{0} - \sum_{i \neq j} \min_{R}(a_{i} x_{i})) / a_{j}$. We have that
this upper bound is non-redundant if and only if
$\lfloor e_{j} \rfloor < ub_{j}$, that is,
$\lfloor e_{j} \rfloor + 1 \leq ub_{j}$, or equivalently,
$e_{j} < ub_{j}$, i.e., $a_{j} e_{j} < a_{j} ub_{j}$. By
expanding the definition of $e_{j}$, this can be rewritten as
$$a_{0} - \sum_{i \neq j} \min_{R}(a_{i} x_{i}) < a_{j} ub_{j}\ ,$$
which holds if and only if
$$
\begin{array}{rcl}
  a_{0} & < & a_{j} ub_{j} + \sum_{i \neq j} \min_{R}(a_{i} x_{i})       \medskip\\
  a_{0} & < & a_{j} ub_{j} - a_{j} lb_{j} + \sum_{i} \min_{R}(a_{i} x_{i}) \medskip\\
  a_{0} & < & | a_{j}| (ub_{j} - lb_{j}) + \sum_{i} \min_{R}(a_{i} x_{i}) \medskip\\
     0 & < & - a_{0} + | a_{j} | (ub_{j} - lb_{j}) + \sum_{i} \min_{R}(a_{i} x_{i})\ . \medskip\\
\end{array}
$$
The case $a_{j} < 0$ is analogous.
By Lemma
\ref{lemma:propagations}, we have $C$ and $R$
propagate $\lceil e_{j} \rceil \bleq x_{j}$, where
$e_{j} = (a_{0} - \sum_{i \neq j} \min_{R}(a_{i} x_{i})) / a_{j}$. We have that
this lower bound is non-redundant if and only if
$\lceil e_{j} \rceil > lb_{j}$, that is,
$\lceil e_{j} \rceil - 1 \geq lb_{j}$, or equivalently,
$e_{j} > lb_{j}$, i.e., $a_{j} e_{j} < a_{j} lb_{j}$. By
expanding the definition of $e_{j}$ this can be rewritten as
$$a_{0} - \sum_{i \neq j} \min_{R}(a_{i} x_{i}) < a_{j} lb_{j}\ ,$$
which holds if and only if
$$
\begin{array}{rcl}
  a_{0} & < & a_{j} lb_{j} + \sum_{i \neq j} \min_{R}(a_{i} x_{i})       \medskip\\
  a_{0} & < & a_{j} lb_{j} - a_{j} ub_{j} + \sum_{i} \min_{R}(a_{i} x_{i}) \medskip\\
  a_{0} & < & | a_{j} | (ub_{j} - lb_{j}) + \sum_{i} \min_{R}(a_{i} x_{i}) \medskip\\
     0 & < & - a_{0} + | a_{j} | (ub_{j} - lb_{j}) + \sum_{i} \min_{R}(a_{i} x_{i})\ . \medskip\\
\end{array}
$$

\bigskip

As regards the second claim, following the previous result, we have
that constraint $C$ propagates a non-redundant bound if and only if
there exists a variable $x_{j}$ such that
  $$|a_{j}| \,(ub_{j} - lb_{j}) > a_{0} - \sum_{i} \min_{R}(a_{i} x_{i})\ ,$$
  and this happens if and only if
$$\max_{j}\, (\;|a_{j}| \,(ub_{j} - lb_{j})\;) > a_{0} - \sum_{i} \min_{R}(a_{i} x_{i})\ .$$  \comment{\qed}


\end{proof}

Using the same notation as in the statement of Lemma
\ref{lemma:filtering}, given a constraint $C$ we define $F'_C$ as
$-a_0 + \max_j\,(\; |a_j| \,(ub_j-lb_j)\;) + \sum_{i} \min_{R}(a_{i}
x_{i})$. By Lemma \ref{lemma:filtering}, constraint $C$ propagates a
non-redundant bound \emph{if, and only if}, $F'_C > 0$. For the sake
of efficiency, the filter $F_C$ of $C$ is actually an upper
approximation of $F'_C$, i.e. $F_{C} \geq F'_C$, so $C$ can only
propagate if $F_C>0$.

To preserve the property that $F_{C} \geq F'_C$, the filters need to
be updated when new bounds are pushed onto the stack (and each update
needs to be undone when popped, for which other data structures
exist). Namely, assume a fresh lower bound $k\bleq x$ is pushed onto
the stack. Let the previous lower bound for $x$ be $k'\bleq x$. Note
that $k' < k$. For each pair $(I_C,a)$ in the positive occurs list of
$x$, using $I_C$ we access the filter $F_C$ and increase it by
$|a(k-k')|$. This accounts for the update of the term
$\sum_{i} \min_{R}(a_{i} x_{i})$ in the definition of $F'_C$, as
$a > 0$ and
$$\min_{k\;\bleq\; x}(a x) = a k = a (k-k') + ak' = |a (k-k')| + \min_{k'\;\bleq\; x}(a x)\ .$$
On the other hand, for efficiency reasons, the term $\max_j\,(\; |a_j| \,(ub_j-lb_j)\;)$ in the definition of $F'_C$ is \emph{not} updated. Since this term decreases as new bounds
are added, we get the inequality $F_{C} \geq F'_{C}$.

Only when $F_C$ becomes positive the constraint $C$ is visited, as it
may propagate some new bound. To avoid too much precision loss, after
each time a constraint $C$ is visited, $F_C$ is reset to the exact
value $F'_C$.

Finally, if a fresh upper bound $x\bleq k$ is pushed onto the stack,
exactly the same is done, but using the negative occurs list and the
previous upper bound $x\bleq k'$ for $x$. Note that $k < k'$, and
if $a < 0$ then
$$\min_{x\;\bleq\; k}(a x) = a k = a (k-k') + ak' = (-a) (k'-k) + ak' = |a (k-k')| + \min_{x\;\bleq\; k'}(a x)\ .$$

\medskip



\medskip

\subsection{Early backjumps}

A significant amount of time in the hybrid resolution and cut-based
conflict analysis of Section \ref{cut-based} is spent on determining
whether an \emph{Early Backjump} can be applied. In our current
implementation, when we want to find the lowest decision level at
which a certain constraint $C$ propagates (a non-redundant bound), we
take each of the variables that occur in $C$ and go over the history
of previous lower and upper bounds in the stack (using the
aforementioned field $pos$). The heights in the stack of these bounds
indicate the decision levels at which $C$ could have propagated
earlier.

The worst-case scenario in this process is when the constraint turns
out not to propagate at any decision level, because all the
potentially propagating decision levels have been examined but the
outcome of the computation is useless. Fortunately, in some situations
it is possible to see in advance that the constraint cannot propagate,
and hence precious time can be saved. Namely, this is the case when
the constraint has been obtained by applying an eliminating cut
inference between two constraints whose only common variable is the
one that is eliminated.

To prove this result formally, first we need the following lemma,
which states that simplifying a constraint by dividing with a common
divisor of the left-hand side and rounding down does not increase
propagation power.

\begin{lemma}
  \label{lemma:simplification-does-not-improve-propagation}
  Let $C_{1}$ be a constraint of the form
  $c a_{1} x_{1} + ... + c a_{n} x_{n} \cleq c k + d$ with
  $0 \leq d < c$, and let $C_{2}$ be the constraint
  $a_{1} x_{1} + ... + a_{n} x_{n} \cleq k$. If $C_{1}$ does not
  propagate any non-redundant bound on a variable $x_{j}$ then neither does $C_{2}$.
\end{lemma}

\begin{proof}

  For any variable $x_{i}$, let $lb_i \bleq x_i$ and $x_i \bleq ub_i$
  be the current lower and upper bounds for $x_i$, respectively. Let
  $A$ be the set of all these bounds.

  Let us assume $C_{1}$ does not propagate any non-redundant bound on $x_{j}$. Then by
  Lemma \ref{lemma:filtering} and since $0 \leq d < c$,
  $$-c k - d + |c\, a_{j}|\,(ub_{j} - lb_{j}) + \sum_{i} \min_{A}(c a_{i} x_{i}) \leq 0$$
  $$c \big(-k + |a_{j}|\,(ub_{j} - lb_{j}) + \sum_{i} \min_{A}(a_{i} x_{i})\big) \leq d$$
  $$c \big(-k + |a_{j}|\,(ub_{j} - lb_{j}) + \sum_{i} \min_{A}(a_{i} x_{i}) \big) < c$$
  $$-k + |a_{j}|\,(ub_{j} - lb_{j}) + \sum_{i} \min_{A}(a_{i} x_{i}) < 1$$
  $$-k + |a_{j}|\,(ub_{j} - lb_{j}) + \sum_{i} \min_{A}(a_{i} x_{i}) \leq 0\ ,$$
  which implies that $C_{2}$ does not propagate any non-redundant bound on $x_{j}$.\comment{\qed}
\end{proof}

Finally, the following lemma ensures that, if the only common variable
is the one that is eliminated, an eliminating cut does not improve the
propagation power.

\begin{lemma}
  \label{lemma:clash}
  Let $C_{1}$ be a constraint of the form
  $a x + a_{1} y_{1} + ... + a_{n} y_{n} \cleq a_{0}$ and $C_{2}$ of
  the form $-b x + b_{1} z_{1} + ... + b_{m} z_{m} \cleq b_{0}$, where
  $a, b > 0$ and the variables $y_{i}$, $z_{j}$ are different
  pairwise. Let $C_{3}$ be the result of applying a cut inference
  eliminating $x$:

  $$C_{3}: b a_{1} y_{1} + ... + b a_{n} y_{n}   +   a b_{1} z_{1} + ... + a b_{m} z_{m}  \cleq   b a_{0} + a b_{0}\ .$$
  Then $C_{3}$ propagates nothing more than $C_{1}$ and $C_{2}$ do.
\end{lemma}

\begin{proof}

  For any variable $v \in \{ x, y_{i}, z_{j}\}$, let $lb(v) \bleq v$
  and $v \bleq ub(v)$ be the current lower and upper bounds for $v$,
  respectively. Let $A$ be the set of all these bounds.

  First of all, by virtue of Lemma
  \ref{lemma:simplification-does-not-improve-propagation} we notice
  that simplifying the cut by dividing with a common factor of the
  left-hand side and rounding down does not increase propagation
  power.

  Now let us assume that $C_{1}$ and $C_{2}$ have been propagated
  exhaustively. As $C_{1}$ does not propagate any non-redundant bound
  on variable $y_{k}$ $(1 \leq k \leq n)$, by Lemma \ref{lemma:filtering}
  $$|a_{k}|\,(ub(y_{k}) - lb(y_{k})) + \min_{A}(a x) + \sum_{i=1}^{n} \min_{A}(a_{i} y_{i}) \leq a_{0}$$
  $$|b a_{k}|\,(ub(y_{k}) - lb(y_{k})) + \min_{A}(a b x) + \sum_{i=1}^{n} \min_{A}(b a_{i} y_{i}) \leq b a_{0} \ .$$

  As $C_{2}$ does not propagate any non-redundant bound
  on variable $x$, by Lemma \ref{lemma:filtering}
  $$|(-b)|\,(ub(x) - lb(x)) + \min_{A}((-b) x) + \sum_{j=1}^{m} \min_{A}(b_{j} z_{j}) \leq b_{0}$$
  $$a b\,(ub(x) - lb(x)) + \min_{A}(a (-b) x) + \sum_{j=1}^{m} \min_{A}(a b_{j} z_{j}) \leq a b_{0}$$
  $$a b\,(ub(x) - lb(x)) - \max_{A}(a b x) + \sum_{j=1}^{m} \min_{A}(a b_{j} z_{j}) \leq a b_{0}\ .$$
  By adding both inequalities:
$$|b a_{k}|\,(ub(y_{k}) - lb(y_{k})) + \min_{A}(a b x) + \sum_{i=1}^{n} \min_{A}(b a_{i} y_{i}) + $$
$$a b \,(ub(x) - lb(x)) - \max_{A}(a b x) + \sum_{j=1}^{m} \min_{A}(a b_{j} z_{j}) \leq b a_{0} + a b_{0}\ ,$$
which simplifies into
$$|b a_{k}|\,(ub(y_{k}) - lb(y_{k})) + \sum_{i=1}^{n} \min_{A}(b a_{i} y_{i}) + \sum_{j=1}^{m} \min_{A}(a b_{j} z_{j}) \leq b a_{0} + a b_{0}$$
as
$\min_{A}(a b x) - \max_{A}(a b x) = ab (\min_{A}(x) - \max_{A}(x)) =
ab (lb(x) - ub(x))$. This shows that $C_{3}$ does not propagate any
non-redundant bound on variable $y_{k}$. The proof of the analogous result
for $z_{l}$ $(1 \leq l \leq m)$ is symmetric.\comment{\qed}
\end{proof}

To apply this result in conflict analysis, let $C_{1}$ be the current
conflicting constraint, and $C_{2}$ be the reason constraint of the
bound $B$ of the conflicting set that is topmost in the stack. Since
we have not been able to apply early backjump so far, $C_{1}$ does not
propagate anything at any previous decision level. And since $C_{2}$
is a constraint of the integer program and before each decision
all bounds are exhaustively propagated, $C_{2}$ cannot propagate
anything at any previous decision level either. Therefore, by Lemma
\ref{lemma:clash}, if $C_{1}$ and $C_{2}$ only have the variable $x$
of bound $B$ in common, and a cut eliminating $x$ exists between
$C_{1}$ and $C_{2}$, then early backjump will not be applicable to the
resulting constraint.

\medskip





\subsection{Clauses}

Clauses are common in ILP problems. For example,
\emph{set covering} constraints, which are of the form
$x_1 +\ldots +x_n \cgeq 1$, are a particular case of clauses where all
literals are positive. As a reference, 50 out of the 240 instances of
the Benchmark Set and 199 out of the 1065 instances of the Collection
Set of the MIPLIB Mixed Integer Programming Library 2017 \footnote{See
  \url{https://miplib.zib.de}.} (roughly 20 \% in both cases) contain
set covering constraints.

Hence there is a potential gain in giving clauses a particular
treatment. For this reason, in our implementation they have a
specialised implementation, with no difference with what could be
found in current SAT solvers. This allows, for example, a more
memory-efficient storage and a faster propagation, thanks to the use
of watch lists instead of occurs lists. Moreover, some clause-specific
constraint simplification techniques can be applied, such as the
so-called lemma shortening introduced in \MiniSAT \cite{MiniSAT},
which has turned out to be essential in the state of the art of SAT
solving.

In particular, \emph{binary clauses} are implemented as edges of the
so-called \emph{binary graph} \cite{Aspvall1979121}. As it is common
in SAT solvers, our implementation uses this graph not only for
propagating even faster than with watch lists, but also for detecting
equivalent literals, which are used to simplify the problem.

Apart from the edges coming from (explicit) binary clauses, we
simulate that the binary graph also contains what we call
\emph{implicit binary clauses}. These are binary clauses that are
consequences of constraints in the integer program, but which are not
explicitly added to the graph. For example, a \emph{set packing}
constraint of the form $x_1 +\ldots +x_n \leq 1$ implies binary
clauses $\bar{x_{i}} \lor \bar{x_{j}}$, where $1 \leq i < j \leq n$.
In general, let us assume we are given an assignment $A$ and we have a
constraint $a_1 x_1+\cdots+ a_n x_n \cleq a_0$ such that $x_{i}$,
$x_{j}$ are Boolean variables satisfying
$1 \cleq x_{i} \not \in A, 1 \cleq x_{j} \not \in A$ and $a_{i}, a_{j} > 0$. Then
the constraint implies (together with $A$) the binary clause
$\bar{x_{i}} \lor \bar{x_{j}}$ if $a_1 x_1+\cdots+ a_n x_n \cleq a_0$
is false in $A \cup \{1 \cleq x_{i}, 1 \cleq x_{j} \}$. By Lemma
\ref{lemma:conflicts}, this happens if and only if
$$a_{i} + a_{j} > a_{0} - \sum_{k=1}^{n} \min_{A}(a_{k} x_{k})\ .$$
In practice, at the beginning of the search, once all bounds have been
exhaustively propagated and before making any decision, all
constraints are examined, and for each literal we store a list of the
identifiers of those constraints that imply a binary clause with that
literal. These lists of constraint identifiers are used in the
algorithms that involve the binary graph to recompute on demand
the binary clauses and thus simulate the corresponding edges.

These three different implementations of constraints (general
constraints, clauses, binary clauses) coexist in the implementation. As a
consequence, the stack of bounds of the assignment has three
integers that point to the constraint, clause and binary clause,
respectively, that was propagated last, and specialised procedures for
propagation are called when these pointers move forward towards the
top of the stack. If a conflict is found, the conflicting constraint
as well as the reason constraints (be they general constraints,
clauses or binary clauses) are transformed into a common
representation in order to carry out the analysis of the conflict.

\medskip

\subsection{Decision heuristics}

As in SAT solvers, our current heuristics for selecting the variable
of the next decision bound is based on recent activity: the variable
with the highest activity score is picked. To that end, a priority
queue of variables, ordered by activity score, is maintained. The
score of a variable $x$ is increased each time a bound containing $x$
appears in the conflicting set during conflict analysis, and to reward
\emph{recent} activity, the amount of increment grows geometrically at
each conflict.

Once a variable $x$ is picked, one has to decide the actual decision
bound: whether it is lower or upper, and how to divide the interval
between the current lower bound $l$ and the current upper bound $u$.
Several strategies have been implemented, among others (in what
follows, let $m$ be the domain middle value $(l+u)/2$):

\begin{enumerate}[label=(\arabic*)]

\item The domain of $x$ is reduced to $[m+1, u]$.
\item The domain of $x$ is reduced to $[u, u]$.
\item The domain of $x$ is reduced to $[l, m]$.
\item The domain of $x$ is reduced to $[l, l]$.

\end{enumerate}

The next two strategies attempt to steer the search towards minimising
the cost in a `first-success' manner. Let $v$ be the best value
for $x$ in $\{l, u\}$ so as to minimise the objective function.

\begin{enumerate}[label=(\arabic*)]
  \setcounter{enumi}{4}

\item The domain of $x$ is reduced to $[l, m]$ if $v = l$, otherwise to $[m+1, u]$.
\item The domain of $x$ is reduced to $[v, v]$.

\end{enumerate}

The following three strategies are inspired by the last-phase polarity
heuristic from SAT \cite{Pipatsrisawat2010JAR} and are aimed at
finding a first solution quickly (which helps to prune the search tree
dramatically). Let $v$ be the last value that $x$ was assigned to.

\begin{enumerate}[label=(\arabic*)]
  \setcounter{enumi}{6}

\item The domain of $x$ is reduced to $[l, m]$ if $v \leq m$, otherwise to $[m+1, u]$.
\item The domain of $x$ is reduced to $[l, l]$ if $v \leq m$, otherwise to $[u, u]$.

\item Assume that $v \in [l, u]$. If $v = l$ or $v = u$, then the
  domain of $x$ is reduced to $[l, l]$ or $[u, u]$, respectively.
  Otherwise, if $v$ is closer to $l$ than to $u$, then the domain of
  $x$ is reduced to $[l, v]$, otherwise to $[v,u]$.
\end{enumerate}

The next strategies are similar to the ones just described, but using
other assignments as a reference.

\begin{enumerate}[label=(\arabic*)]
  \setcounter{enumi}{9}

\item Like (7), (8), (9), but using the value of $x$ in the last
  solution.

\item Like (7), (8), (9), but using the value of $x$ in an initial
  solution provided by the user.

\end{enumerate}

The user can specify an order between these strategies, in such a way
that if a strategy cannot be applied (for instance, in (7), (8), (9)
if variable $x$ has not been assigned yet, or in (9) if
$v \not\in [l, u]$), then the next strategy in the order is attempted.

\section{Experiments}
\label{experiments}

This section is devoted to the experimental evaluation of
\warning{\IntSat}. After describing the competing solvers in Section
\ref{solvers} and explaining how the benchmarks have been obtained in
Section \ref{instances}, the results of this evaluation are shown in
Section \ref{results}. Finally Section
\ref{sec:experiment-conclusions} presents some conclusions that can be
drawn from the evaluation.

All experiments reported next were carried out on a standard 3.00 GHz
8-core Intel i7-9700 desktop with 16 Gb of RAM.
A binary of our basic implementation as well as all benchmarks can be
downloaded from \cite{IntSatSoftwareDownload2020}, so that the
interested reader can reproduce and verify the results reported here.

\subsection{Solvers}
\label{solvers}

In this experimental evaluation we compare our implementation with the
newest versions of the following solvers:

\begin{enumerate}

\item \Gurobi (v.9.5.1)\\
(\url{www.gurobi.com/})


\item \CPLEX (v.22.1.0.0)\\
(\url{www.ibm.com/products/ilog-cplex-optimization-studio})


\item \SCIP (v.8.0.0)\\
(\url{www.scipopt.org/})


\item \GLPK (v.5.0)\\
(\url{www.gnu.org/software/glpk/})

\end{enumerate}

\CPLEX and \Gurobi are well-known commercial ILP solvers, while \GLPK
and \SCIP are non-commercial. It has to be pointed out that the
technology behind these solvers is extremely mature, after decades of
improvements: according to \cite{Bixby2021}, between 1990 and 2020
they have seen a 3.7 million times speedup from
\emph{machine-independent} algorithmic improvements only (that is, a
factor of 2.1 \emph{per year})! Rather than applying a single
one-size-fits-all method, these solvers resort to a large variety of
techniques, including, e.g., specialised cuts (Gomory, knapsack, flow
and GUB covers, MIR, clique, zero-half, mod-k, network, submip, etc.),
heuristics (rounding, RINS, solution improvement, feasibility pump,
diving, etc.) and branching strategies (pseudo cost branching, strong
branching, reliability branching, etc.). In what follows we will argue
in favour of including the techniques presented in this article in
this toolkit.

Due to their prevalence and great success in practice, in this
evaluation we focus on comparing with Operations Research ILP solvers
based on the simplex algorithm and branch-and-cut. However, there are
other tools that can handle ILP with different approaches, which have
not been included here mainly for performance reasons. For instance,
SAT Modulo Theories (SMT) \cite{Nieuwenhuisetal2006JACM} solvers such
as \textsf{Mathsat5} \cite{mathsat5}, \textsf{Yices} \cite{Yices},
\textsf{Z3} \cite{Z3} or our own \textsf{Barcelogic} solver
\cite{Bofilletal2008CAV} are aimed at problems consisting of Boolean
combinations of integer linear constraints, being ILP the particular
case of \emph{conjunctions} of constraints. SMT solvers specialize in
efficiently handling this arbitrary Boolean structure, while their
\emph{theory solver} component, the one that precisely handles
conjunctions of constraints (our goal here), is not as sophisticated
as in simplex-based systems. As a consequence, the performance of SMT
solvers on ILP instances is in general orders of magnitude worse than
\CPLEX or \Gurobi, \warning{and for this reason they are
not included in these experiments.}
%

Concerning SAT and Lazy Clause Generation (LCG)
\cite{DBLP:conf/cp/OhrimenkoSC07}, from our own work (see among many
others \cite{encodeOrPropagate}), we also know too well that solvers
that (lazily) encode integer linear constraints into SAT can be
competitive as long as problems are mostly Boolean, without a heavy
numerical/optimisation component. Also CSP solvers such as
\textsf{Sugar} \cite{Sugar} or \textsf{Gecode}
\cite{DBLP:journals/constraints/SchulteT13}, which heavily focus on
their rich constraint language, are in general very far from
commercial Operations Research solvers on hard ILP optimisation
problems. \warning{Because of that, these tools have not been considered in our
experimental comparison either.}



Finally, another solver very close to our methods is \CutSat
\cite{JovanovicDeMouraJAR2013}, which also attempts to generalise CDCL
from SAT to ILP. As argued in Section \ref{rounding-problem}, the
approach that \CutSat follows to work around the rounding problem
poses important limitations, which translate in practice into a rather
poor performance when compared to state-of-the-art ILP
solvers, typically orders of magnitude slower \cite{Nieuwenhuis2014CP}.
Moreover, the current implementation can
only handle feasibility problems and has no optimisation features.

\subsection{Instances}
\label{instances}

The instances used in the experiments were taken from the MIPLIB Mixed
Integer Programming Library. The latest edition of the library (MIPLIB
2017, available at \url{https://miplib.zib.de}) organises benchmarks
into two sets: the \emph{Benchmark Set}, which contains 240 instances
that are solvable by (the union of) today's codes and were chosen
`subject to various constraints regarding solvability and numerical
stability'; and the much larger \emph{Collection Set}, which
represents a diverse selection of instances \cite{MIPLIB2017}. See the
MIPLIB website for statistics of each instance, including among others
the number of continuous, integer and binary variables, the number of
non-zeroes in the coefficient matrix, and the number of constraints in
different classes: set covering, set packing, set partitioning,
cardinality, knapsack, etc.

From these two sets of instances, we picked those that:

\begin{enumerate}

\item do not contain continuous variables, and

\item have lower and upper bounds for all variables, and

\item do not contain constraints or objective functions with
  fractional coefficients with many (four or more) decimal digits, and

\item contain at least one integer non-binary variable, and

\item are known to be infeasible or have a feasible solution,
  according to MIPLIB 2017.

\end{enumerate}

Restrictions 1-3 in the above list are due to the limitations of our
algorithms. In particular, to ensure that all coefficients are
integer, the constraints and objective functions with fractional
coefficients are multiplied by appropriate powers of $10$.
Moreover, instances with only binary variables were discarded, as our
goal here are non-binary problems; to deal with binary linear
programming, a specialised implementation would perform much better
(see Section \ref{sec:binary-ilp}). Finally, instances with an
`open' status at MIPLIB 2017 (that is, which are still not known to
have a feasible solution or not) were not included after confirming
that, within a reasonable amount of time, no solver could produce an
answer for them.

After applying this filtering, we obtained 29 benchmarks from the
Benchmark Set, and 40 from the Collection Set. We also added to our
selection the three instances from MIPLIB 2010 that were used in our
earlier work in \cite{Nieuwenhuis2014CP} and that have been removed
from the current 2017 edition of the library. Altogether we compiled a
test suite consisting of 72 benchmarks.

In order to present the results in a coherent way, in what follows
these instances will be classified depending on whether they admit a
feasible solution or not, and whether they are feasibility or
optimisation problems. Hence, out of the 72 instances of the benchmark
suite, it turns out that 6 are infeasible:
\texttt{cryptanalysiskb128n5obj14}, \texttt{no-ip-64999},
\texttt{no-ip-65059}, \texttt{fhnw-sq3}, \texttt{neos-3211096-shag}
and \texttt{neos859080}. In fact, all of these but \texttt{neos859080}
are actually feasibility problems, i.e., there is no objective
function.
The remaining 66 feasible instances of the test suite contain both
feasibility and optimisation problems. As regards the former there are
6 feasibility instances, namely:
\texttt{cryptanalysiskb128n5obj16},
\texttt{neos-3004026-krka},
\texttt{fhnw-sq2},
\texttt{lectsched-1},
\texttt{lectsched-2} and
\texttt{lectsched-3}.
The rest of the benchmark suite consists of 60 feasible optimisation
instances.

\subsection{Results}
\label{results}

The \warning{time limit} of all executions was set to 1 hour of wall-clock time.
However, it is important to highlight that our implementation of \warning{\IntSat}
(as well
as \GLPK and \SCIP) is sequential and only uses one core, while on the
other hand \CPLEX and \Gurobi are run in parallel mode. Therefore,
they may use (and often do use) all of the eight cores that are
available in the computer used for the experiments.

\subsubsection{Infeasible instances}
\label{sec:infeasible-instances}

\begin{table}
\tbl{Infeasible instances.\bigskip}
{\begin{tabular}{l|rrr|rrrrrr|}
& \multicolumn{3}{c|}{\textbf{Problem statistics}}
& \multicolumn{6}{c|}{\textbf{Time}}
  \\
 &\#cons & \#vars & \#ints  & \textsf{grb} & \textsf{cpx} & \textsf{scip} & \textsf{glpk} &  \textsf{isr} & \textsf{isc} \\
  \hline
\texttt{no-ip-64999} &     2547 &     2232 &       45 &   \b{1.7}       &        364.0       &        625.0    &        TO        &           TO &           TO       \\
\texttt{no-ip-65059} &     2547 &     2232 &       45 &   \b{2.6}       &         30.6       &        145.3    &        TO        &           TO &           TO       \\
\texttt{cryp..bj14}  &    98021 &    48950 &     1120 &    3036.9       &           TO       &        TO       &        TO        &       3031.9 &    \b{444.9}       \\
\texttt{fhnw-sq3}    &      167 &     2450 &       49 &        TO       &           TO       &        TO       &        TO        &           TO &           TO       \\
\texttt{neos..shag}  &    10187 &     4379 &       90 &        TO       &           TO       &        TO       &        TO        &           TO &           TO       \\
\texttt{neos859080}  &      164 &      160 &       80 &       0.2       &       \b{0.1}      &       1.4       &   \b{0.1}        &           TO &           TO       \\

 \end{tabular}
}
\label{tab:infeasible}
\end{table}

Table \ref{tab:infeasible} summarizes the results on infeasible
benchmarks. There is a row for each instance\footnote{Due to space
  limitations, the names of some of the instances are only partially
  displayed. However, the shrunk name is enough to identify them,
  e.g., using the search engine of the MIPLIB website.}. As regards
columns, the first three describe size properties of the instance:
number of constraints, (total) number of variables and number of
(non-binary) integer variables. Recall that the number of continuous
variables is always zero here. The rest of the columns show the time
in seconds that each solver takes to prove infeasibility:
\begin{enumerate}

\item \textsf{grb} stands for \Gurobi;

\item \textsf{cpx} stands for \CPLEX;

\item \textsf{scip} stands for \SCIP;

\item \textsf{glpk} stands for \GLPK;

\item \warning{\textsf{isr}} (short for `\IntSat Resolution') stands for our core \warning{\IntSat} algorithm with the
  implementation of the procedure
  \textsf{Conflict\_analysis\_Backjump\_Learn} as described in Section
  \ref{set-based}.

\item \warning{\textsf{isc}} (short for `\IntSat Cuts') is similar to the previous solver, but with
  \textsf{Conflict\_analysis\_Backjump\_Learn} implemented as in
  Section \ref{cut-based}.

\end{enumerate}
The timing TO stands for time out. The fastest solver for each problem
(if any) is highlighted in bold face.

First of all, we remark the difficulty of most of these instances.
Even for a tool like \CPLEX, for half of the benchmarks infeasibility
cannot be proved within the time limit of 1 hour; and over the 6
instances, \Gurobi times out twice. Having said that, it appears that
our techniques are not especially appropriate for this kind of
problems: infeasibility can be proved only for a single instance.
However, it is worth highlighting the complementarity of the
techniques: for this particular instance
\texttt{cryptanalysiskb128n5obj14}, which turns out to be a difficult
problem even for \Gurobi, the overall best solver, our solver
\warning{\textsf{isc}} performs comparatively well.

\subsubsection{Feasible feasibility instances}
\label{sec:feasible-feasibility-instances}

\begin{table}
\tbl{Feasible feasibility instances.\bigskip}
{\begin{tabular}{l|rrr|rrrrrr|}
 &  \multicolumn{3}{c|}{\textbf{Problem statistics}}
 &  \multicolumn{6}{c|}{\textbf{Time}}
  \\
  & \#cons  &  \#vars  &  \#ints   &  \textsf{grb}  &  \textsf{cpx}  &  \textsf{scip}  &  \textsf{glpk}  &   \warning{\textsf{isr}}  &  \warning{\textsf{isc}} \\
  \hline
\texttt{crypt..obj16} &  98021 &   48950  &   1120  &   TO   &   TO   &   TO     &  TO  &   576.1  &\b{38.6} \\
\texttt{neos..krka}   & 12545  &   17030  &    130  &   7.3  &   TO   &   266.4  &  TO  & \b{0.1}  &\b{0.1}  \\
\texttt{fhnw-sq2}     &    91  &     650  &     25  &   TO   &   TO   &   TO     &  TO  &   TO     &   TO    \\
\texttt{lectsched-1}  & 50108  &   28718  &    482  &   6.4  &  13.1  &   450.0  &  TO  & \b{0.3}  &\b{0.3}  \\
\texttt{lectsched-2}  & 30738  &   17656  &    369  &   0.5  &   2.2  &     6.0  &  TO  & \b{0.1}  &\b{0.1}  \\
\texttt{lectsched-3}  & 45262  &   25776  &    457  &   1.1  &   4.1  &   199.2  &  TO  & \b{0.2}  &\b{0.2} \\
 \end{tabular}
 }
\label{tab:feasible0}
\end{table}

Table \ref{tab:feasible0} shows the results obtained on feasible
feasibility benchmarks. The format of the table is as in Table
\ref{tab:infeasible}, but here the last six columns indicate the time
in seconds that each solver takes to prove feasibility (that is, to
find a solution, given that in these instances there is no objective
function).

As can be seen in the table, for these feasibility problems our
\warning{\IntSat} solvers turn out to perform reasonably well.
Any of the two uniformly
performs better than any other solver, even \CPLEX and \Gurobi.
Although admittedly the \texttt{lectsched-*} instances are rather
easy, the other three are not so (e.g., \CPLEX times out on all of
them, \emph{even with the eight available cores}). Moreover, in the
particular case of the instance \texttt{cryptanalysiskb128n5obj16},
in fact our \warning{\IntSat} solvers
are the only ones that could find a solution. It
is worth noting that \warning{\textsf{isc}} solved this instance in roughly 38
seconds.

\subsubsection{Feasible optimisation instances}
\label{sec:feasible-optimization-instances}

In this subsection we report the results on feasible optimisation
instances. Here the focus will be on the ability of the solvers
\emph{to find good solutions quickly},
rather than their strength in proving optimality. We do so motivated
on practical grounds.
Indeed, for many real-life instances it is simply impossible
to certify optimality in the allotted time.
As noted in \cite{JOHNSON198039}, this can consume a huge fraction of the long overall running time.
This is natural, as optimal solutions can be
discovered using good heuristic guidance and may involve only
exploring a small part of the space of solutions, while proving
optimality, on the other hand, involves reasoning over the whole
search space. For this precise reason, MILP solvers based on the
simplex method and branch-and-cut actually only look for new solutions
until they consider they are `close enough' to the optimum, according to their MILP gap tolerance.

In consequence, in what follows feasible optimisation instances are
divided into two groups, depending on whether or not an optimal
solution could be \emph{found} in the executions with the different
solvers. In order to identify solutions as optimal without requiring
solvers to effectively prove optimality, we use the optimal value of
the objective function as reported in MIPLIB \footnote{Particular
  cases are instances \texttt{neos-4360552-sangro},
  \texttt{neos-3426132-dieze} and \texttt{proteindesign121hz512p19},
  which according to MIPLIB are still open and therefore their optimal
  values are not available. For these problems we have used the
  objective values of the best known solutions as a proxy.}.

\begin{table}
\tbl{Feasible optimisation instances, time for finding the optimal solution.\bigskip}
{\begin{tabular}{l|r@{\hskip 0.05in}r@{\hskip 0.05in}r|r@{\hskip 0.1in}r@{\hskip 0.1in}r@{\hskip 0.1in}r@{\hskip 0.1in}r@{\hskip 0.1in}r@{\hskip 0.05in}|}
 &  \multicolumn{3}{c|}{\textbf{Problem statistics}}
 &  \multicolumn{6}{c|}{\textbf{Time}}
  \\[0.01cm]
  & \#cons  &  \#vars  &  \#ints   &  \textsf{grb}  &  \textsf{cpx}  &  \textsf{scip}  &  \textsf{glpk}  &   \warning{\textsf{isr}}  &  \warning{\textsf{isc}} \\[0.01cm]
 \hline
\texttt{30n20b8}        &     576  &    18380  &      62  & 1.5      & 1.7       & 88.7      & TO      & 35.0       & \b{0.7}   \\[0.01cm]
\texttt{comp07-2idx}    &   21235  &    17264  &     109  & \b{27.4} & 333.6     & TO        & TO      & TO         & 38.5      \\[0.01cm]
\texttt{comp08-2idx}    &   12536  &   11554   &      67  & \b{ 5.3} & 918.3     & 657.0     & TO      & TO         & TO        \\[0.01cm]
\texttt{gt2}            &      29  &      188  &     164  &\b{0.0}   & \b{0.0}   & \b{0.0}   & TO      & 15.6       & \b{0.0}   \\[0.01cm]
\texttt{hypo..id-k1}    &    5195  &     2602  &       1  & 7.5      & 12.8      & 15.4      & 25.1    & \b{0.3}    & \b{0.3}   \\[0.01cm]
\texttt{k1mushroom}     &   16419  &     8211  &       1  & 38.1     & 330.3     & 942.8     & TO      & \b{5.6}    & 6.1       \\[0.01cm]
\texttt{lect..4-obj}    &   14163  &    7901   &     236  & 0.5      & 2.0       & 3.5       & TO      & \b{0.4}    & 0.5       \\[0.01cm]
\texttt{lect..5-obj}    &   38884  &   21805   &     416  & TO       & TO        & TO        & TO      & \b{129.7}  & 240.8     \\[0.01cm]
\texttt{manna81}        &    6480  &    3321   &    3303  & \b{0.1}  & 0.2       & 0.3       & 10.6    & TO         & TO        \\[0.01cm]
\texttt{mzzv11}         &    9499  &    10240  &     251  & \b{3.5}  & 7.7       & 181.0     & TO      & 41.2       & 51.3      \\[0.01cm]
\texttt{mzzv42z}        &   10460  &    11717  &     235  & \b{1.8}  & 2.0       & 208.7     & TO      & 205.4      & 80.1      \\[0.01cm]
\texttt{neos-1223462}   &    5925  &     5495  &     315  & \b{0.3}  & 0.5       & 79.9      & TO      & TO         & 2.8       \\[0.01cm]
\texttt{neos..4597}     &    3486  &     3395  &     245  & \b{0.0}  & 0.2       & 0.8       & TO      & 7.1        & 0.3       \\[0.01cm]
\texttt{neos-1354092}   &    3135  &    13702  &     420  &\b{1461.5}& TO        & TO        & TO      & TO         & TO        \\[0.01cm]
\texttt{neos-1582420}   &   10180  &    10100  &     100  & 4.0      & \b{1.1}   & 30.5      & TO      & TO         & TO        \\[0.01cm]
\texttt{neos..loue}     &    3705  &     3255  &    3255  & \b{30.3} & TO        & 1748.9    & TO      & TO         & TO        \\[0.01cm]
\texttt{neos..nubu}     &    4725  &     8644  &    8644  & 0.5      & \b{0.4}   & 5.6       & 234.0   & TO         & TO        \\[0.01cm]
\texttt{neos..apure}    &     169  &  100000   &  100000  & 15.6     & \b{6.3}   & 157.3     & 76.4    & TO         & 3538.9    \\[0.01cm]
\texttt{neos..awhea}    &     479  &     2375  &    1900  & \b{1.0}  & 1.5       & 3.8       & TO      & TO         & TO        \\[0.01cm]
\texttt{neos..dieze}    &     570  &   11550   &    8800  & \b{90.6} & TO        & TO        & TO      & TO         & TO        \\[0.01cm]
\texttt{neos..gauja}    &     220  &    2310   &    1890  & \b{8.2}  & 3328.0    & TO        & TO      & TO         & TO        \\[0.01cm]
\texttt{neos..gaula}    &     200  &     2090  &    1900  & \b{2.0}  & 2.5       & TO        & TO      & TO         & TO        \\[0.01cm]
\texttt{neos..glina}    &   44206  &   57196   &   22924  & \b{0.2}  & 0.5       & 9.1       & 0.4     & 0.5        & 0.5       \\[0.01cm]
\texttt{neos..wannon}   &   85865  &   31728   &   31248  & 4.1      & 16.1      & 113.5     & 1933.7  & \b{1.2}    & 3.2       \\[0.01cm]
\texttt{neos-555001}    &    3474  &     3855  &      73  & \b{0.1}  & 1.1       & 0.9       & TO      & TO         & 41.0      \\[0.01cm]
\texttt{neos-555343}    &    3326  &     3815  &      15  & \b{3.5}  & 61.6      & TO        & TO      & TO         & 710.1     \\[0.01cm]
\texttt{neos-555424}    &    2676  &     3815  &      15  & \b{1.4}  & 45.6      & 521.6     & TO      & TO         & 93.7      \\[0.01cm]
\texttt{neos-555884}    &    4331  &     3815  &      15  & \b{0.6}  & 3257.6    & 102.1     & TO      & TO         & TO        \\[0.01cm]
\texttt{neos-662469}    &    1085  &    18235  &     328  & \b{28.8} & 73.7	 & 2475.2    & TO      & TO         & TO        \\[0.01cm]
\texttt{neos-686190}    &    3664  &    3660   &      60  & 10.5     & \b{2.5}   & 74.6      & 114.3   & TO         & TO        \\[0.01cm]
\texttt{neos-950242}    &   34224  &     5760  &     240  & 0.2      & 11.9      & 126.8     & 35.6    & \b{0.1}    & \b{0.1}   \\[0.01cm]
\texttt{neos16}         &    1018  &      377  &      41  & \b{0.6}  & 14.0      & 83.6      & TO      & 130.2      & 81.5      \\[0.01cm]
\texttt{neos8}          &   46324  &    23228  &       4  &    0.5   & \b{0.3}   & 8.2       & 7.1     & 1.0        & 0.9       \\[0.01cm]
\texttt{nurse..ium04}   &    8668  &   29667   &      54  & \b{11.3} & 3434.6    & TO        & TO      & TO         & TO        \\[0.01cm]
\texttt{nurse..den09}   &    4872  &   11650   &      20  & \b{7.7}  & 416.9     & 72.8      & TO      & TO         & TO        \\[0.01cm]
\texttt{nurse..ate03}   &    5032  &   11690   &      20  & \b{9.9}  & 270.8     & 160.2     & TO      & TO         & TO        \\[0.01cm]
\texttt{nurse..nt02}    &    3522  &    10250  &      20  & 3.84     & \b{2.92}  & 28.4      & TO      & TO         & TO        \\[0.01cm]
\texttt{prot..z512p9}   &     301  &   159145  &      91  & \b{127.5}& 3386.5    & TO        & TO      & TO         & TO        \\[0.01cm]
\texttt{prot..b11p9}    &     254  &  132672   &      78  & TO       & 3417.9    & TO        & TO      & TO         & \b{294.3} \\[0.01cm]
\texttt{prot..x11p8}    &     254  &   127326  &      78  & \b{354.2}& TO        & TO        & TO      & TO         & 1581.1    \\[0.01cm]
\texttt{roco..011000}   &    1667  &     4456  &     136  & 382.8    & \b{228.7} & 1859.1    & TO      & TO         & TO        \\[0.01cm]
\texttt{roco..001000}   &    1293  &     3117  &     124  & 231.3    & \b{75.2}  & 337.0     & TO      & TO         & TO        \\[0.01cm]
\texttt{roco..010100}   &    4010  &   12321   &     166  & \b{354.7}& 2220.8    & TO        & TO      & 716.2      & TO        \\[0.01cm]
\texttt{roco..011100}   &    2367  &     6491  &     166  & \b{10.4} & 924.9     & TO        & TO      & TO         & TO        \\[0.01cm]
\texttt{sp98ir}         &    1531  &     1680  &     688  & \b{8.2}  & 8.4       & 28.1      & 1190.7  & TO         & TO        \\[0.01cm]
\texttt{splice1k1}      &    6505  &    3253   &       1  & 255.6    & 320.3     & 1155.5    & TO      & \b{14.4}   & 18.8      \\[0.01cm]
\texttt{supp..case33}   &   20489  &    20203  &     101  & 25.3     & \b{12.0}  & 191.4     & TO      & TO         & 213.5     \\[0.01cm]
\texttt{neos..avoca}    &    1570  &     2368  &    2325  &\b{298.5} & TO        & TO        & TO      & TO         & TO        \\[0.01cm]
\texttt{neos..ticino}   &     308  &     4688  &    3809  &\b{28.0}  & TO        & TO        & TO      & TO         & TO        \\[0.01cm]
\end{tabular}
}
\label{tab:feasible1}
\end{table}

Namely, Table \ref{tab:feasible1} displays the results on the 49
instances for which at least one solver could find an optimal solution
within the time limit of 1 hour. The format of the table is as in
Tables \ref{tab:infeasible} and \ref{tab:feasible0}, but now the last
six columns indicate the time in seconds that each solver takes to
find an optimal solution. Here the timing TO means that the solver
timed out before finding an optimal solution (although it may have
obtained other worse solutions, which is not represented in the table
for the sake of succinctness).

\begin{table}
\tbl{Feasible optimisation instances, cost of best solution after different run times.\bigskip}
{\begin{tabular}{@{\hskip -0.05in}l|@{\hskip -0.0in}r@{\hskip 0.05in}r@{\hskip 0.05in}r|@{\hskip -0.0in}r@{\hskip 0.08in}r@{\hskip 0.08in}r@{\hskip 0.08in}r@{\hskip 0.08in}r@{\hskip 0.08in}r@{\hskip 0.05in}|c|}
 &  \multicolumn{3}{c|}{\textbf{Problem statistics}}
 &  \multicolumn{6}{c|}{\textbf{Value}}
 &  \multicolumn{1}{c|}{\textbf{\warning{Time}}}
  \\
  & \#cons  &  \#vars  &  \#ints   &  \textsf{grb}  &  \textsf{cpx}  &  \textsf{scip}  &  \textsf{glpk}  &   \warning{\textsf{isr}}  &  \warning{\textsf{isc}} & \textbf{\warning{lapse}}    \\
  \hline

\texttt{supp..case1}  &  59997 &   29999 &    200 &   ---          &   ---       &   ---        &   ---       &   ---          & \b{296362}        & \warning{\textrm{ 1 min}} \\
                      &        &         &        &   ---          &   ---       &   ---        &   ---       &   ---          & \b{296362}        & \warning{\textrm{ 5 min}} \\
                      &        &         &        &   ---          &   ---       &   ---        &   ---       &   ---          & \b{293902}        & \warning{\textrm{15 min}} \\
                      &        &         &        &   ---          &   ---       &   ---        &   ---       &   ---          & \b{293848}        & \warning{\textrm{30 min}} \\
                      &        &         &        &   ---          &   ---       &   ---        &   ---       &   ---          & \b{293848}        & \warning{\textrm{60 min}} \\[0.20cm]

\texttt{neos..sovi}   &   7244 &    4318 &    179 &   ---          &   ---       &   ---        &   ---       &   ---          &   ---             & \warning{\textrm{ 1 min}} \\
                      &        &         &        &   ---          &   ---       &   ---        &   ---       &   ---          &   ---             & \warning{\textrm{ 5 min}} \\
                      &        &         &        &   ---          &   ---       &   ---        &   ---       &   ---          &   ---             & \warning{\textrm{15 min}} \\
                      &        &         &        &   ---          &   ---       &   ---        &   ---       & \b{180185}     &   180245         & \warning{\textrm{30 min}} \\
                      &        &         &        &   ---          &   ---       &   ---        &   ---       & \b{180165}     &   180185          & \warning{\textrm{60 min}} \\[0.20cm]

\texttt{roc..111000}  &  10776 &    8619 &    187 & \b{36689}      &   37623     &   38773      &   ---       &   82087        &   ---           & \warning{\textrm{ 1 min}} \\
                      &        &         &        & \b{36247}      &   36379     &   38773      &   ---       &   82087        &   ---           & \warning{\textrm{ 5 min}} \\
                      &        &         &        & \b{35770}      &   36284     &   38773      &   ---       &   82087        &   87188         & \warning{\textrm{15 min}} \\
                      &        &         &        & \b{35488}      &   36221     &   38773      &   ---       &   82087        &   87188         & \warning{\textrm{30 min}} \\
                      &        &         &        & \b{35469}      &   35588     &   38773      &   ---       &   82087        &   87188         & \warning{\textrm{60 min}} \\[0.20cm]

\texttt{comp12-2idx}  &  16803 &   11863 &     43 & \b{666}        &   ---       &   ---        &   ---       &   1035         &   825           & \warning{\textrm{ 1 min}} \\
                      &        &         &        & \b{351}        &   455       &   ---        &   782       &   870          &   825           & \warning{\textrm{ 5 min}} \\
                      &        &         &        & \b{350}        &   401       &   934        &   782       &   784          &   769           & \warning{\textrm{15 min}} \\
                      &        &         &        & \b{350}        &   383       &   695        &   753       &   784          &   678           & \warning{\textrm{30 min}} \\
                      &        &         &        & \b{350}        &   364       &   682        &   658       &   784          &   678           & \warning{\textrm{60 min}} \\[0.20cm]

\texttt{supp..case19} &  10713 & 1429098 & 117806 &   ---          &   ---       &   ---        &   ---       &   ---          &   ---           & \warning{\textrm{ 1 min}} \\
                      &        &         &        &   ---          &   ---       &   ---        &   ---       &   ---          &   ---           & \warning{\textrm{ 5 min}} \\
                      &        &         &        &   ---          &   ---       &   ---        &   ---       &   ---          &   ---           & \warning{\textrm{15 min}} \\
                      &        &         &        &   ---          &   ---       &   ---        &   ---       &   ---          &   ---           & \warning{\textrm{30 min}} \\
                      &        &         &        &   ---          &   ---       &   ---        &   ---       &   ---          &   ---           & \warning{\textrm{60 min}} \\[0.20cm]

\texttt{ns1854840}    & 143616 &  135754 &    474 &  309200       &   ---       &   427200   &   ---       &   126000      & \b{65200}        & \warning{\textrm{ 1 min}} \\
                      &        &         &        &  309200       &   ---       &   427200   &   ---       &   126000      & \b{54400}        & \warning{\textrm{ 5 min}} \\
                      &        &         &        &  309200       &   ---       &   427200   &   ---       &   125800      & \b{26800}        & \warning{\textrm{15 min}} \\
                      &        &         &        &   27800       &   ---       &   427200   &   ---       &    97000      & \b{26800}        & \warning{\textrm{30 min}} \\
                      &        &         &        &\b{25000}      &   ---       &   427200   &   ---       &    94000      &    26800         & \warning{\textrm{60 min}} \\[0.20cm]

\texttt{roco..010001} &   4636 &   16741 &    187 &\b{36314}       &    40328    &    85756     &   ---       &   65656        &   65742         & \warning{\textrm{ 1 min}} \\
                      &        &         &        &\b{35532}       &    36566    &    43177     &   ---       &   65562        &   61523         & \warning{\textrm{ 5 min}} \\
                      &        &         &        &\b{35485}       &    35515    &    42166     &   ---       &   60697        &   59654         & \warning{\textrm{15 min}} \\
                      &        &         &        &   35479        & \b{34633}   &    41795     &   ---       &   60697        &   57719         & \warning{\textrm{30 min}} \\
                      &        &         &        &   35122        & \b{34436}   &    41795     &   ---       &   60697        &   55983         & \warning{\textrm{60 min}} \\[0.20cm]
\texttt{neos..sangro} &  46012 &   10272 &    576 &   ---          &   ---       &   ---        &   ---       &   ---          &   \b{-6}   & \warning{\textrm{ 1 min}} \\
                      &        &         &        &   ---          &   ---       &   ---        &   ---       &   ---          &   \b{-6}   & \warning{\textrm{ 5 min}} \\
                      &        &         &        &    -6          &   ---       &   ---        &   ---       &    -6          &   \b{-7}   & \warning{\textrm{15 min}} \\
                      &        &         &        &    -6          &   ---       &   ---        &   ---       &    -6          &   \b{-7}   & \warning{\textrm{30 min}} \\
                      &        &         &        &    -6          &   ---       &   ---        &   ---       &    -6          &   \b{-7}   & \warning{\textrm{60 min}} \\[0.20cm]

\texttt{nurse..int03} &  14062 &   34248 &     78 &\b{4662}        &   ---       &   8080       &   ---       &   10534        &    5652    & \warning{\textrm{ 1 min}} \\
                      &        &         &        &\b{ 137}        &   334       &   8080       &   ---       &    9534        &    3064    & \warning{\textrm{ 5 min}} \\
                      &        &         &        &\b{ 119}        &   232       &   8080       &   ---       &    6925        &    2779    & \warning{\textrm{15 min}} \\
                      &        &         &        &\b{ 117}        &   171       &   7266       &   ---       &    6925        &    2277    & \warning{\textrm{30 min}} \\
                      &        &         &        &\b{ 116}        &   123       &   1056       &   ---       &    4797        &    1966    & \warning{\textrm{60 min}} \\[0.20cm]

\texttt{comp21-2idx}  &  14038 &   10863 &     71 &   118          &\b{110}      &   699        &   ---       &   319          &    210     & \warning{\textrm{ 1 min}} \\
                      &        &         &        &\b{ 83}         &   106       &   250        &   250       &   319          &    190     & \warning{\textrm{ 5 min}} \\
                      &        &         &        &\b{ 83}         &   102       &   156        &   211       &   319          &    190     & \warning{\textrm{15 min}} \\
                      &        &         &        &\b{ 83}         &    93       &   150        &   211       &   319          &    183     & \warning{\textrm{30 min}} \\
                      &        &         &        &\b{ 76}         &    84       &   148        &   211       &   319          &    163     & \warning{\textrm{60 min}} \\[0.20cm]

\texttt{prot..12p19}  &    301 & 2589931 &     91 &\b{3412}        &   ---       &   ---        &   ---       &   ---          &    ---     & \warning{\textrm{ 1 min}} \\
                      &        &         &        &\b{3412}        &   ---       &   ---        &   ---       &   ---          &    ---     & \warning{\textrm{ 5 min}} \\
                      &        &         &        &\b{3408}        &   ---       &   ---        &   ---       &   ---          &    ---     & \warning{\textrm{15 min}} \\
                      &        &         &        &\b{3406}        &   ---       &   ---        &   ---       &   ---          &    3602    & \warning{\textrm{30 min}} \\
                      &        &         &        &\b{3406}        &   ---       &   ---        &   ---       &   ---          &    3475    & \warning{\textrm{60 min}} \\[0.20cm]
\end{tabular}}
\label{tab:feasible2}
\end{table}

Table \ref{tab:feasible2} shows the results on
the remaining 11 instances. Given that none of the solvers could find
an optimal solution, this table is different in that it does not
display timings but how the objective function value of the best solution
evolved along time. The format is similar to that in previous tables,
with the difference that now each problem spans five rows, indicating
the value of the best solution that could be found within the first 1
minute, 5 minutes, 15 minutes, 30 minutes and 60 minutes,
respectively. \warning{Moreover,
  there is an additional final column with these time lapses to help
  reading the results.} A dash means that the \warning{time lapse
  passed} without discovering any solution. For each time lapse,
the solver with the best objective value over all is highlighted in bold face.

First of all, let us compare our \warning{\IntSat} solvers between themselves.
Out of 49
instances, in Table \ref{tab:feasible1} there are 7 instances in which
\warning{\textsf{isr}} gives better results than \warning{\textsf{isc}}, and among these,
1 instance in which \warning{\textsf{isc}} actually times out. On the other
hand, there are 15 instances in which \warning{\textsf{isc}} is better than
\warning{\textsf{isr}}, and among these, 9 instances in which \warning{\textsf{isr}} times
out. As regards Table \ref{tab:feasible2}, \warning{\textsf{isr}} is clearly
superior to \warning{\textsf{isc}} in instance \texttt{rococoC12-111000}, while
the reverse occurs in instances \texttt{supportcase1},
\texttt{comp12-2idx}, \texttt{comp21-2idx}, \texttt{rococoC12-010001},
\texttt{ns1854840}, \texttt{proteindesign121hz512p19}
and \texttt{nursesched-medium-hint03}.

Altogether, although the conflict analysis with cuts and early
backjumps as described in Section \ref{cut-based} usually pays off,
there is a significant number of instances on which a simpler conflict
analysis yields better results. This complementarity suggests that,
given that they are now run sequentially, a portfolio approach running
both variants in parallel would be in order.

Now let us contrast our solvers with \SCIP and \GLPK. To simplify the
comparison, for each instance we will consider the best of
\texttt{res} and \texttt{cut}, and the best of \SCIP and
\GLPK. Regarding Table \ref{tab:feasible1}, we see that the best of
our solvers was superior to the best solver of \SCIP and \GLPK in 20
instances, and the other way around in 19 instances. As for Table
\ref{tab:feasible2}, our solvers were better in instances
\texttt{supportcase1}, \texttt{neos-3214367-sovi},
\texttt{proteindesign121hz512p19},
\texttt{neos-4360552-sangro}
and \texttt{ns1854840},
and the other way around for instances \texttt{rococoC12-010001} and
\texttt{rococoC12-111000}. In short, we see our solvers tend to
perform slightly better than \SCIP and \GLPK.

Finally, let us compare our solvers with \CPLEX and \Gurobi. Again,
for the sake of simplicity, for each instance we will consider the
best of \texttt{res} and \texttt{cut}, and the best of \CPLEX and
\Gurobi. We see that in Table \ref{tab:feasible1} the best of our
solvers was superior to the best solver of \CPLEX and \Gurobi in 9
instances, and the other way around in 39 instances. As for Tables
\ref{tab:feasible2}, our solvers were better
in instances \texttt{supportcase1}, \texttt{neos-3214367-sovi} and
\texttt{ns1854840}, and the other way around for instances
\texttt{comp12-2idx}, \texttt{rococoC12-010001}, \texttt{neos-4360552-sangro},
\texttt{nursesched-medium-hint03}, \texttt{comp21-2idx}
and
\texttt{rococoC12-111000}. Although on the whole admittedly \CPLEX and
\Gurobi are superior, there is about a 20\% of the instances for which
our solvers obtain better results, which is a non-negligible
percentage. It has also to be reminded that, while
\warning{\textsf{isr}} and \warning{\textsf{isc}} are sequential and
only use one core, \CPLEX and \Gurobi are parallel and often use all
eight available cores.

\subsection{Conclusions of the Experiments}
\label{sec:experiment-conclusions}

Altogether, the results of the experiments in Sections
\ref{sec:infeasible-instances},
\ref{sec:feasible-feasibility-instances} and
\ref{sec:feasible-optimization-instances} indicate that (both) our
\warning{\IntSat} algorithms can be helpful on optimisation problems so as to find first
solutions of relatively good quality. This is even more the case for
feasibility problems, which are more amenable to our methods as
optimisation is not natively supported. Our results thus confirm the
well-known fact that depth-first search, which is the search strategy
that CDCL solvers implement, is particularly appropriate for finding
solutions fast. In fact, a typical search strategy in ILP solvers
consists in first performing depth-first search in order to get a
feasible solution to enable cost-based pruning in branch-and-bound,
and then apply best-first search, so as to give more emphasis to the
quality of the solutions. Taking this into account we consider that,
in the diverse toolkit of techniques that ILP solvers implement,
\warning{\IntSat} could provide an edge when feasibility is prioritised over
optimality, or in the first stages of the search.

\section{Future Work}
\label{further}

\warning{A large number of further ideas around \warning{\IntSat} and
its implementation are yet to be explored, among which we sketch the
most relevant ones in this section.}

\subsection{Unboundedness}

The current implementation assumes that for each variable there is an
initial lower bound and an upper bound. This is used in the proof of
Theorem \ref{theorem:termination-soundness-completeness} to ensure the
termination of the core \warning{\IntSat} algorithm. Although it is common that this
condition holds in real-life applications, some instances do have
unbounded variables. In these cases non-termination may manifest
itself in, e.g., bound propagation. For example, consider the set of
constraints consisting of $C_1\!\!: x - y \cleq 0$ and
$C_2\!\!: -x + y + 1 \cleq 0$, which clearly does not have any
solution. If we decide bound $0 \bleq x$, constraint $C_1$ propagates
$0 \bleq y$, then $C_2$ propagates $1 \bleq x$, then constraint $C_1$
propagates $1 \bleq y$, and so on indefinitely, resulting into an
endless chain of propagations. Hence, even bound propagation is not
guaranteed to terminate for unbounded infeasible problems.

In theory, any ILP can be converted into an equivalent fully bounded
one \cite{Schrijver86}. Unfortunately, these bounds turn out to be too
large to be useful in practice.

A pragmatic solution to handle unbounded variables is to introduce a
fresh auxiliary variable $z$ with lower bound $0 \bleq z$, and for
each variable $x$ without lower bound add the constraint $-z \bleq x$,
and similarly if it has no upper bound add $x \bleq z$. Then one can
re-run the algorithm with successively larger upper bounds
$z \bleq ub$ for $z$, thus guaranteeing completeness for finding
(optimal) solutions.

An alternative strategy for handling unbounded variables is to apply
the so-called \emph{bounding transformation} presented in
\cite{DBLP:conf/cade/Bromberger18}. This allows reducing any problem
to an equisatisfiable bounded one via a Mixed-Echelon-Hermite
transformation composed with a double-bounded reduction.

\subsection{Restarts and Constraint Database Management}

From the practical point of view, our current implementation
mimics several ideas from CDCL SAT solving without having tested them
thoroughly.

For instance, like SAT solvers our basic implementation applies periodic
\emph{restarts}. We currently follow a policy that triggers a restart
when the number of conflicts reaches a threshold. The user can specify
if this threshold is determined according to the Luby sequence
\cite{luby1993} or to an inner-outer geometric series
\cite{DBLP:journals/jsat/Biere08}. Moreover, each of these strategies
has a number of parameters with a significant effect on performance and
whose values are still to be tuned.

Another aspect that requires further investigation is the
\emph{cleanup} policy for the constraint database; at each cleanup, we
remove all non-initial constraints with more than two monomials and
activity counter equal to 0. This activity counter is increased each
time the constraint is a conflicting or reason constraint at conflict
analysis, and is divided by 2 at each cleanup. Cleanups are done
periodically in such a way that the constraint database grows rather
slowly over time. The policy that is currently implemented triggers a
cleanup after the number of new learnt constraints reaches a
threshold, or the memory used by learnt constraints exceeds a space
limit. We think there is still room for improvement in adjusting the
parameters of this strategy based on experimental grounds.
Furthermore, other cleanup strategies can be devised, e.g., by
generalising the heuristics based on the \emph{literal block distance}
\cite{Audemard2009} that are used in state-of-the-art SAT solvers with
great success.

Finally, modern SAT solvers heavily apply pre- and in-processing
techniques \cite{EenBiere2005SAT,DBLP:conf/sat/FazekasBS19} to keep the constraint
database small but strong. These techniques could be worth applying
also in the context of CDCL solvers for ILP.

\subsection{Conflict analysis}

Several ideas can be explored for conflict analysis. Instead of taking
the extreme position of always trying to apply early backjump (as in
Section \ref{cut-based}) or never (as in Section \ref{set-based}), one
can, for instance, attempt to do an early backjump with the
intermediate conflicting constraint $C$ only if it is false in the
current stack, or promising (e.g., short) according to some heuristic.
Other conflict analysis algorithms mixing resolution with reason and
conflicting sets and cuts with reason and conflicting constraints can
also be designed.

In another line of investigation, the quality of the backjumps and the
strength of the reason sets could be improved by doing some more work:
e.g., instead of using the pre-stored reason sets, one can re-compute
them on the fly during conflict analysis with the aim of maximising
the length of the backjump. One can also do a bit of search during
conflict analysis, e.g., by trying to remove non-topmost bounds and do
cuts with these, with the aim of finding good early backjump cuts.

\medskip




\medskip


\subsection{Binary ILP and Mixed ILP}
\label{sec:binary-ilp}

Binary ILP is a particular case of ILP, and as such the algorithms
presented here are directly applicable. However, given the prevalence
of this kind of problems, it is worth to specialise the design and
implementation of the data structures and subprocedures, e.g. in the
propagation mechanism. Preliminary experiments with a binary ILP
solver prototype \cite{DBLP:journals/access/NieuwenhuisORR21} already
show a significant speed-up in running time in comparison with the
general basic implementation.

In the opposite direction, it also needs to be worked out how to apply
the proposed algorithms in order to solve mixed ILP (MILP) instances,
i.e., where not all variables are subject to integrality. For
instance, one could decide on the integer variables as it is done now,
and at any desired point, run an LP solver to optimise the values for
the rational variables. The inclusion of lower bounding techniques,
well-known from modern MILP solvers, needs to be considered as well.

\section{Related Work and Conclusions}
\label{conc}

The success of SAT solving has spurred a number of
research projects that attempt to extend CDCL-related techniques to
ILP and MILP. Some of these come from the traditional MILP community.
For example, in \cite{DBLP:journals/disopt/Achterberg07} conflict
analysis is generalised to mixed integer programming thanks to special
heuristics for branch-and-cut. These techniques are implemented in
\SCIP, one of the solvers used in the experiments in Section
\ref{experiments}.

Other works come from the SAT area, as the aforementioned \CutSat
presented in \cite{JovanovicDeMouraJAR2013}, which has later been
refined and extended in \cite{DBLP:journals/jsc/BrombergerSW20}.
Unlike in these works, our cut-based reasoning does not replace but is
performed \emph{in parallel} to resolution-based reasoning with reason
and conflicting sets.

In this same direction, the idea of applying reason and conflicting
sets is not only reminiscent to the conflict analysis of SAT, but also
of \emph{SAT Modulo Theories} (SMT)
\cite{Nieuwenhuisetal2006JACM,BarrettHandbookSMT2009} for the theory
of linear arithmetic, with the main difference, among others, that
here new ILP constraints are obtained by cut inferences, normalised
and learned, and not only new Boolean clauses that are disjunctions of
literals representing bounds (usually only those that occur in the
input formula). Other SAT/SMT related work, but for rational
arithmetic is
\cite{McMillanKS2009CAV,KorovinVoronkov2011CADE,Cotton2010Formats}.

As outlined in Section \ref{cut-based}, it is also worth mentioning
that there may be some possible theoretical and practical consequences
of the fact that our algorithm's underlying cutting planes proof
system is stronger than CDCL's resolution proof system: could we
outperform SAT solvers on certain SAT problems (e.g.,
pigeon-hole-like) for which no short resolution proofs exist? A
similar question applies to the current SMT solvers, which are based
on resolution as well \cite{NieuwenhuisIJCAR2012}.

It seems unlikely that for ILP or MILP solving one single technique
can dominate the others; the best solvers will probably continue
combining different methods from a large toolbox, which perhaps will
also include this work at some point. \warning{Still, \warning{\IntSat} already
appears to be the first alternative method for ILP that, without
resorting to the simplex algorithm or LP relaxations, turns out to be
competitive on hard optimisation problems.} We expect that, give its
large potential for enhancement, this work will trigger further
research activity, in particular along the lines sketched in
Section~\ref{further}.

\section*{Conflict of interest}

The authors declare that they have no conflict of interest.

\section*{Funding}

All authors are supported by grant PID2021-122830OB-C43, funded by MCIN/ AEI/10.13039/501100011033 and by “ERDF: A way of making Europe”. The Version of Record of this manuscript has been published and is available in Optimization Methods and Software 27 Sep 2023 \url{https://www.tandfonline.com/doi/abs/10.1080/10556788.2023.2246167}

\bibliographystyle{tfs}
\bibliography{../../bibNew}

\appendix
\section{Proofs of Lemmas on Bound Propagation}
\label{app:bound-propagation}

In this appendix we detail all the proofs of the lemmas stated in
Section \ref{bound-propagation}.

\bigskip

\begin{proof}[Proof of Lemma \ref{lemma:conflicts}]
  For the first claim, if $a_{i} > 0$ then $a_i x_i$ is
  increasing in $x_{i}$, so the minimum value takes place at the least
  value that $x_{i}$ can take, that is $lb_{i}$, and at that point
  $a_i x_i$ evaluates to $a_i\cdot lb_i$. Note that, by convention, if
  $lb_{i} = -\infty$ then $a_i\cdot lb_i = -\infty$ as $a_{i} > 0$.

  Similarly, if $a_{i} < 0$ then $a_i x_i$ is decreasing in $x_{i}$,
  so the minimum value takes place at the largest value that $x_{i}$
  can take, that is $ub_{i}$, and at that point $a_i x_i$ evaluates to
  $a_i\cdot ub_i$. Note that, by convention, if $ub_{i} = +\infty$
  then $a_i\cdot ub_i = -\infty$ as $a_{i} < 0$.


\medskip

  As regards the second claim, we observe that the constraint is false if and
  only if $\min_{A}(a_1 x_1 + \ldots + a_nx_n) > a_{0}$. And since $A$
  is a set of bounds,
  $\min_{A}(a_1 x_1 + \ldots + a_nx_n) = \min_{A}(a_1 x_1) + \ldots +
  \min_{A}(a_nx_n)$.
  \comment{\qed}

\end{proof}

\bigskip

\begin{proof}[Proof of Lemma \ref{lemma:propagations}]
  Let us consider the case $a_{j} > 0$. Let us assume that there
  exists a solution $\mathit{sol}$ to $\{ C \} \cup R$ with
  $\mathit{sol}(x_{j}) = v > \lfloor e_{j} \rfloor$
  and we will get a contradiction. If
  $v > \lfloor e_{j} \rfloor$ then
  $v \geq \lfloor e_{j} \rfloor + 1$, which implies
  $v > e_{j}$, that is, $a_{j}\; v > a_{j} e_{j}$. Expanding the definition of $e_{j}$ we get
  $$ a_{j} v > a_{0} - \sum_{i \neq j} \min_{R}(a_{i} x_{i})$$
  or equivalently
  $$ a_{j} v + \sum_{i \neq j} \min_{R}(a_{i} x_{i}) > a_{0}\ ,$$
  which by Lemma \ref{lemma:conflicts} implies that
  $\{ C \} \cup R \cup \{v \bleq x_{j}, x_{j} \bleq v \} $ has no
  solution, a contradiction.

  The case $a_{j} < 0$ is similar. If there exists a solution
  $\mathit{sol}$ to $\{ C \} \cup R$ with
  $\mathit{sol}(x_{j}) = v < \lceil e_{j} \rceil$, then
  $v \leq \lceil e_{j} \rceil - 1$, which implies $v < e_{j}$, that
  is, $a_{j}\; v > a_{j} e_{j}$. The proof then follows in the same
  way as in the case $a_{j} > 0$. \comment{\qed}
\end{proof}

\bigskip

\begin{proof}[Proof of Lemma \ref{no-rounding}]
  { By Lemma \ref{lemma:conflicts}, $C_{1}$ is false in
    $R \cup \{ x_{j} \preceq e_{j}\}$ if and only if
    $\sum_{i} \min_{R}(a_{i} x_{i}) > a_{0}$. But
    $\min_{R \cup \{ x_{j} \preceq e_{j}\}}(a_{j} x_{j}) = a_{j}
    e_{j}$ as $a_{j} < 0$. So we have
    $a_{j} e_{j} + \sum_{i \neq j} \min_{R}(a_{i} x_{i}) > a_{0}$.
  By expanding the definition of $e_{j}$ and multiplying at both sides
  by $b_{j} > 0$, we get
  $$a_{j} b_{0} - a_{j} \sum_{i \neq j} \min_{R}(b_{i} x_{i}) + b_{j} \sum_{i \neq j} \min_{R}(a_{i} x_{i}) > a_{0} b_{j}\ .$$
  But
  $$ - a_{j} \sum_{i \neq j} \min_{R}(b_{i} x_{i}) + b_{j} \sum_{i \neq j} \min_{R}(a_{i} x_{i}) =$$
  $$ = \sum_{i \neq j} \min_{R}(- a_{j} b_{i} x_{i}) +  \sum_{i \neq j} \min_{R}(b_{j} a_{i} x_{i}) \leq \sum_{i \neq j} \min_{R}((- a_{j} b_{i} + b_{j} a_{i}) x_{i})\ .$$
  Altogether
  $$\sum_{i \neq j} \min_{R}((- a_{j} b_{i} + b_{j} a_{i}) x_{i}) > a_{0} b_{j} - a_{j} b_{0}$$ and by Lemma \ref{lemma:conflicts} we have proved that $C_{3}$ is false in $R$.
}
\comment{\qed}
\end{proof}

\section{Proof of Termination, Soundness and Completeness of \IntSat}
\label{app:theorem:termination-soundness-completeness}

In this appendix we prove in full detail Theorem
\ref{theorem:termination-soundness-completeness}, which states the
termination, soundness and completeness of the core \warning{\IntSat}
algorithm.

To that end some definitions are
required. Given a stack of bounds $A$ and a bound $B \in A$, we write
the height of $B$ in the stack as $\height_{A}(B)$.
Moreover, if a stack of bounds $A$ is of the form
$A_{0}\, \dec_{1}\, A_{1}\, \cdots\, A_{n-1}\, \dec_{n}\, A_{n}$,
where $\dec_{1}$, ..., $\dec_{n}$ are all the decision bounds in $A$,
then we define the $0$-th \emph{decision level} as $A_{0}$, the $1$st
decision level as $\dec_{1}\, A_{1}$, and in general for $i > 0$, the
$i$-th decision level as $\dec_{i}\, A_{i}$. If a bound $B$ belongs to
the $i$-th decision level, then we say that $B$ \emph{has} decision level
$i$.

We will need the following lemma:

\begin{lemma}
  \label{lemma:invariants}
  Let us assume procedure \textsf{Conflict\_analysis\_Backjump\_Learn}
  is valid.
  Let $S_{0}$ be the finite set of constraints given as input to the
  core \warning{\IntSat} algorithm. Let $S$ be the set of constraints
  and $A$ be the stack of bounds at any step of the execution
  of the algorithm. Then the following properties hold:

  \begin{enumerate}

  \item\label{item:constraint-set-sols}
    $S$ and $S_{0}$ have the same solutions: $S \models S_{0}$ and
    $S_{0} \models S$.

  \item\label{item:stack-no-contradictory} $A$ does not contain
    contradictory bounds.

  \item\label{item:stack-no-redundant}
    For any bounds $B, B' \in A$, if
    $\height_{A}(B) < \height_{A}(B')$ then $B'$ is not redundant with $B$.

  \item\label{item:stack-reason-set} For any non-decision bound
    $B \in A$, its reason set $RS$ is such that: (i) $RS \subseteq A$, (ii)
    $\height_{A}(B') < \height_{A}(B)$ for any $B'\in RS$, and (iii)
    $S_{0} \cup RS \models B$.

  \item\label{item:stack-reason-constraint}
    For any non-decision bound $B \in A$ with reason constraint $RC$,
    it holds that $S_{0} \models RC$.

  \item\label{item:stack-prop}
    If $A$ is of the form
    $A_{0}\, \dec_{1}\, A_{1}\, \cdots\, A_{n-1}\, \dec_{n}\, A_{n}$, where
    $\dec_{1}$, ..., $\dec_{n}$ are all the decision bounds in $A$, then
    $S_{0} \cup \{ \dec_{1}, \ldots, \dec_{i} \} \models A_{i}$ for all $i$ in
    $0...n$.

  \end{enumerate}

\end{lemma}

\begin{proof}
  Given an execution of the algorithm, let us prove the properties by
  induction over the number of steps.


  Let us prove the base case. At the beginning of the execution we
  have that $S = S_{0}$. Therefore property
  \ref{item:constraint-set-sols} holds trivially.
  Moreover, initially we also have that $A$ is empty. So properties
  from \ref{item:stack-no-contradictory}
  to \ref{item:stack-prop}
  hold too.

  Now let us prove the inductive case. Let us see that, if the
  properties hold at any step of the execution of the algorithm, then
  they are true at the next step.

  To begin with, we notice that constraints are only added to $S$ at
  step 4. These constraints are the result of calling procedure
  \textsf{Conflict\_analysis\_Backjump\_Learn}. By hypothesis, any
  such constraint $C$ satisfies that $S \models C$. As
  $S_{0} \models S$, then $S_{0} \models S \cup \{C\}$. And since
  $S \models S_{0}$, we also have that $S \cup \{C\} \models S_{0}$.
  Altogether, property \ref{item:constraint-set-sols} is preserved
  when $S$ is updated.

  As regards the properties involving the stack of bounds, we will use
  that bounds are pushed in the following circumstances:

  \begin{enumerate}

  \item[A.] At step 1, when for a constraint $RC \in S$ and a set of
    bounds $RS \subseteq A$, we have that $RC$ and $RS$ propagate a
    fresh bound $B$.

  \item[B.] At step 2, when we decide a fresh bound $B$.

  \item[C.] At step 4, when we push a fresh bound $B$ obtained from
    the call to procedure
    \textsf{Conflict\_analysis\_Backjump\_Learn}.

  \end{enumerate}

  This case distinction will be used in the proofs of the remaining
  properties that follow.

  Next let us see that the stack of bounds cannot contain redundant or
  contradictory bounds (properties \ref{item:stack-no-contradictory}
  and \ref{item:stack-no-redundant}). Note that in propagations (case
  A) and decisions (case B), only bounds that are fresh are added to
  $A$ by construction, and hence cannot be redundant or contradictory.
  As regards case C, let $A'$ be the stack returned by the call to
  \textsf{Conflict\_analysis\_Backjump\_Learn}. By hypothesis, the
  assignment $A$ can be decomposed as $N\, D\, M$, where $N$ and $M$
  are sequences of bounds and $D$ is a decision bound, and $A'$ is of
  the form $N\, B$, where $B$ is a fresh bound in $N$. By induction
  hypothesis, since $N$ is a prefix of $A$ and $B$ is fresh in $N$,
  when $A$ is replaced by $A'$, we have that properties
  \ref{item:stack-no-contradictory} and \ref{item:stack-no-redundant}
  hold.

  Now let us prove that, after we push a non-decision bound $B$ onto
  the stack, properties \ref{item:stack-reason-set} and
  \ref{item:stack-reason-constraint} hold. We distinguish the
  following cases (note that case B does not apply, as $B$ is not a
  decision):

  \begin{itemize}

  \item[A.] By construction, the reason set $RS$ of $B$ satisfies
    $RS \subseteq A$. This implies that, after pushing $B$, bounds in
    $RS$ will be below $B$ in the stack. Moreover, if $RC \in S$ is
    the reason constraint of $B$, then $\{RC\} \cup RS \models B$ by
    Lemma \ref{lemma:propagations}. As $RC \in S$, we have that
    $S \cup RS \models B$. And since $S_{0} \models S$, finally
    $S_{0} \cup RS \models B$. This proves property
    \ref{item:stack-reason-set}. Moreover, since $RC \in S$ and
    $S_{0} \models S$, we also have that $S_{0} \models RC$, which
    proves property \ref{item:stack-reason-constraint}.

  \item[C.] Let $A'$ be the stack returned by the call to
    \textsf{Conflict\_analysis\_Backjump\_Learn}. By the validity of
    the procedure, the assignment $A$ can be decomposed as
    $N\, D\, M$, where $N$ and $M$ are sequences of bounds and $D$ is
    a decision bound, and $A'$ is of the form $N\, B$, where $B$ is a
    fresh bound in $N$. By hypothesis, $RS \subseteq N$ and
    $S \cup RS \models B$. Therefore, bounds in $RS$ are below $B$ in
    the stack $A'$. And $S_{0} \models S$ and $S \cup RS \models B$
    imply $S_{0} \cup RS \models B$, which proves property
    \ref{item:stack-reason-set}. Finally, if the reason constraint
    $RC$ of $B$ is defined, then by the validity of
    \textsf{Conflict\_analysis\_Backjump\_Learn} $S \models RC$, which
    together with $S_{0} \models S$ implies that $S_{0} \models RC$,
    thus proving property \ref{item:stack-reason-constraint}.

  \end{itemize}

  Now let us prove property \ref{item:stack-prop}. It is enough to
  prove that, whenever a non-decision bound $B$ is pushed onto a stack
  $A$ of the form
  $A_{0}\, \dec_{1}\, A_{1}\, \cdots\, A_{n-1}\, \dec_{n}\, A_{n}$,
  where $\dec_{1}$, ..., $\dec_{n}$ are all the decision bounds in
  $A$, then
  $S_{0} \cup \{ \dec_{1}, \ldots \cup, \dec_{n} \} \models B$. First,
  we notice that by property \ref{item:stack-reason-set}, the reason
  set $RS$ of $B$ is such that $RS \subseteq A$ and
  $S_{0} \cup RS \models B$.
  \warning{Now, by recursively repeating the same reasoning for the previous
  non-decision bounds that there may be in $RS$, we have that the
  bounds in the left-hand side of the implication
  $S_{0} \cup RS \models B$ can be reduced to only decisions and
  $S_{0} \cup \{ \dec_{1}, \ldots, \dec_{n} \} \models B$.}

  To conclude the proof of Lemma \ref{lemma:invariants}, we observe
  that all properties are trivially preserved when bounds are popped
  from the stack of bounds. The only exception is property
  \ref{item:stack-reason-set} (i), i.e., that for any non-decision
  bound $B \in A$, its reason set $RS$ is such that $RS \subseteq A$.
  But since bounds in $RS$ are below $B$ in the stack, before removing
  any element of $RS$ we must have popped $B$ already. \comment{\qed}
\end{proof}

\begin{proof}[Proof of Theorem \ref{theorem:termination-soundness-completeness}]
  First of all, let us prove that the core algorithm terminates. To
  that end,
  extending \cite{Nieuwenhuisetal2006JACM}, 
  we define a well-founded ordering $\succ$ on the states of
  the stack $A$, as follows. For a given $A$, the \emph{size of the
    domain} of $x_{i}$, that is the number of possible values that the
  variable can still take, is $v_i(A) = \upsilon_i - \lambda_i + 1$,
  where $\lambda_i \bleq x_i$ and $x_i \bleq \upsilon_i$ are the
  topmost (and so strongest) lower and upper bounds of $x_{i}$ in $A$.
  The aggregated domain size for all $n$ variables is then
  $v(A) = v_1(A)+\ldots v_n(A)$. For $i\geq 0$ let $A_i$ denote the
  bottom part of $A$, below (and without) the $(i+1)$-th decision.
  \footnote{Here $A_{i}$ has a different meaning from that in the
    definition of decision level, or in property \ref{item:stack-prop}
    of Lemma \ref{lemma:invariants}.}
  We define a stack $A$ to be
  larger (i.e., less advanced, search-wise) than a stack $A'$, written
  $A \succ A'$, if
  $\langle v(A_0),\ldots,v(A_m)\rangle >_{lex} \langle
  v(A'_0),\ldots,v(A'_m) \rangle$, where $m$ is the maximum number of
  decisions the stack can contain. This number is at most
  $\sum_{i=1}^{n} (ub_i - lb_i)$, where $lb_i \bleq x_i$ and
  $x_i \bleq ub_i$ are respectively the lower and the upper bounds of
  the variable $x_{i}$ in $S_{0}$, which are guaranteed to exist by
  the hypothesis of Theorem
  \ref{theorem:termination-soundness-completeness}. Note that, when a
  fresh bound over variable $x_{i}$ is pushed onto the stack,
  the size of the domain of $x_{i}$ is
  decremented from $s > 1$ (otherwise, the bound could not be fresh)
  to $s' < s$; hence at most $ub_i - lb_i$ decisions can be made on
  $x_{i}$.
  The ordering $\succ$ over stacks of bounds is well-founded as an
  infinitely decreasing chain of stacks would lead to an infinitely
  decreasing chain of tuples of natural numbers over the lexicographic
  ordering, which is well known to be well-founded.

  To complete the proof of termination, let us see that at every step
  the algorithm either halts
  or transforms the stack of bounds $A$ into an $A'$ with $A \succ A'$.

  First let us observe that, if a bound $B$ over variable $x_{i}$ is
  fresh in $A$ and is pushed onto $A$, resulting into $A' = AB$, then
  $A \succ A'$. Indeed, if $j$ is the last decision level of $A'$
  (that is, exactly $j$ decisions have been made in $A'$), as $B$ is
  fresh we have that $v_{i}(A_j) > v_{i}(A_j')$ and
  $v_{k}(A_j) = v_{k}(A_j')$ for all $k \not = i$. So
  $v(A_j) > v(A_j')$. And since $A_p = A_p'$ for all $0 \leq p < j$,
  we have $v(A_p) = v(A_p')$. Altogether, $A \succ A'$.

  Now, in the propagation loop (step 1), the previous observation
  ensures that every propagation makes the stack decrease according to
  $\succ$. Hence the loop terminates either because no fresh bounds
  can be propagated, or due to a conflict. If there is no conflict,
  and if no fresh bounds can be decided, the algorithm halts.
  Otherwise a fresh bound is decided. Using again the previous
  observation, we see that the stack decreases according to $\succ$.

  On the other hand, when there is a conflict, the algorithm halts if
  the stack contains no decisions. Otherwise, procedure
  \textsf{Conflict\_analysis\_Backjump\_Learn} is called. By the
  validity of the procedure, this call terminates. Moreover, the stack
  $A$ can be decomposed as $N\, D\, M$, where $N$ and $M$ are
  sequences of bounds and $D$ is a decision bound, and $A'$ is of the
  form $N\, B$, where $B$ is a fresh bound in $N$. Let us assume that
  $D$ is the $(i+1)$-th decision of $A$. Then $A_{i} = N$ and
  $A'_{i} = N\, B$, which implies that $v(A_{i}) > v(A'_{i})$ as $B$
  is fresh in $N$. Furthermore, for all $0 \leq p < i$ we have
  $v(A_p) = v(A_p')$ since $A_p = A_p'$. Altogether finally we see
  that $A \succ A'$, which concludes the proof of termination.

  Now, as regards correctness, we observe that the algorithm halts

  \begin{itemize}

  \item[A.] when a conflict is found and the current stack of bounds contains no
    decision (returning `infeasible'), or

  \item[B.] when there is no conflict
  and no fresh bound can be decided (returning `solution $A$').

  \end{itemize}

  In case A, by property \ref{item:stack-prop} of Lemma
  \ref{lemma:invariants} we have that $S_{0} \models A$, and by
  property \ref{item:constraint-set-sols} that $S \models S_{0}$,
  which together imply that $S \models A$. But, on the other hand, the
  conflict $C \in S$ is such that $\{C\} \cup A$ is infeasible, which
  implies that $S \cup A$ is infeasible. Altogether $S$ is infeasible,
  and so is $S_{0}$ by property \ref{item:constraint-set-sols}.

  In case B, as no fresh bound can be decided, $A$ determines
  a total assignment. And since there is no conflict, $A \models C$ for
  all $C \in S$. Hence, $A$ is a solution to $S$, and by virtue
  of property \ref{item:constraint-set-sols} of Lemma
  \ref{lemma:invariants}, it is a solution to $S_{0}$ as well.\comment{\qed}
\end{proof}

\section{Proofs of validity of \textsf{Conflict\_}\textsf{analysis\_}\textsf{Backjump\_Learn}}
\label{app:validity}

In this appendix we proof Theorems \ref{valid-resolution-based} and
\ref{valid-cut-based}, which ensure that different ways of
implementing the procedure
\textsf{Conflict\_}\textsf{analysis\_}\textsf{Backjump\_Learn} are
valid.

\begin{proof}[Proof of Theorem \ref{valid-resolution-based}]

  To start with, let us prove that the properties that
  $CS \subseteq A$ and that $S\cup CS$ is infeasible are indeed an
  invariant of conflict analysis, by induction on the number of steps.

  At the beginning $CS \subseteq A$ holds by construction. Moreover,
  as $C$ is false in $CS$ we also have that $\{ C \} \cup CS$ is
  infeasible, which together with $C \in S$ implies that $S\cup CS$ is
  infeasible.

  Now let us assume that the property holds, and let us see that it is
  preserved in the next iteration of conflict analysis. If $RS$ is the
  reason set of a non-decision bound $B \in A$, then $RS \subseteq A$
  by property \ref{item:stack-reason-set} of Lemma
  \ref{lemma:invariants}, and therefore the condition $CS \subseteq A$
  from the induction hypothesis implies that
  $(CS\setminus \{B\}) \cup RS \subseteq A$. Let us see now that
  $S\cup (CS\setminus \{B\}) \cup RS$ is infeasible by contradiction.
  If there is a solution $sol$, then in particular $sol$ is a solution
  to $S\cup RS$. Since by properties \ref{item:constraint-set-sols}
  and \ref{item:stack-reason-set} of Lemma \ref{lemma:invariants} we
  have $S \cup RS \models B$, this implies that $sol$ satisfies $B$.
  But by definition $sol$ also satisfies $CS\setminus \{B\}$.
  Altogether, $sol$ is a solution to $S \cup CS$, which is a
  contradiction as $S \cup CS$ is infeasible by induction hypothesis.

  Once we have proved the invariants of conflict analysis, we can
  proceed with the proof of validity of \proc. First, let us show its
  termination. Conflict analysis terminates because at each iteration
  the height in the stack $A$ of the topmost bound of $CS$ is
  decreased, by virtue of property \ref{item:stack-reason-set} of
  Lemma \ref{lemma:invariants}. Note that, if this topmost bound is a
  decision, then the exit condition of the loop is satisfied and
  conflict analysis halts. Finally, the termination of backjumping and
  learning is straightforward.

  Let us prove now that for any constraint $C \in T$ we have
  $S \models C$ (property \ref{def:valid:consequence} in the
  definition of validity). If $T = \emptyset$ there is nothing to
  prove. Otherwise $T = \{ CC\}$, where $CC$ is a constraint
  equivalent to the final $\no{CS}$, viewed as a clause (disjunction)
  of bounds. By the invariants of conflict analysis, $S\cup CS$ is
  infeasible. Therefore, for any $sol$ solution to $S$, we have that
  $sol$ cannot satisfy $CS$ and hence there must be a bound $B \in CS$
  such that $sol \models \no B$. But this implies that $sol$ satisfies
  the clause $\no{CS}$, and so $CC$. Since $sol$ is arbitrary, we have
  that indeed $S \models CC$.



  Let us show now property \ref{def:valid:decomposition} in the
  definition of validity. The stack $A$ can be decomposed as
  $N\, D\, M$, where $N$ and $M$ are sequences of bounds and $D$ is
  the last bound that is popped in backjumping. Note that this bound
  $D$ must be a decision. Therefore $A'$ has to be of the form
  $N\, \no{B_{top}}$. It remains to be seen that $\no{B_{top}}$ is a
  fresh bound in $N$. But if $\no{B_{top}}$ were redundant with some
  bound $B \in N$, then $B_{top}$ would be contradictory with $B$ in
  $A$, contradicting property \ref{item:stack-no-contradictory} of
  Lemma \ref{lemma:invariants}. Similarly, if $\no{B_{top}}$ were
  contradictory with some bound $B \in N$, then $B_{top}$ would be
  redundant with $B$ in $A$, contradicting property
  \ref{item:stack-no-redundant} of Lemma \ref{lemma:invariants}.
  Hence, $\no{B_{top}}$ is indeed fresh in $N$.

  Let us now prove property \ref{def:valid:reason-set} in the
  definition of validity.
  To that end we need to characterise the last popped bound in
  backjumping. Since, by the invariants of conflict analysis, we have
  $CS \setminus \{B_{top}\} \subseteq A$, we can consider the maximum
  decision level in $A$ of the bounds in $CS \setminus \{B_{top}\}$,
  which we will denote by $l$ (with the convention that $l = 0$ if
  $CS = \{B_{top}\}$). Note that $A$ contains at least $l+1$
  decisions, since by definition of $B_{top}$ its decision level is
  strictly greater than $l$. So let us define $D'$ as the $(l+1)$-th
  decision in $A$, and let us prove that $D'$ is the last bound that
  is popped from $A'$ in backjumping. Let us distinguish two cases:

  \begin{itemize}

  \item If $CS = \{B_{top}\}$ then $l = 0$ and, according to the
    exit conditions, backjumping pops bounds from $A'$ until
    there are no decisions left. Hence the last popped bound is the
    first decision, that is the $(l+1)$-th decision, which is $D'$.

  \item Otherwise, if $CS \setminus \{B_{top}\} \not= \emptyset$ we
    observe that, for any bound $B$ in this set, we have that
    $\height_{A}(B) < \height_{A}(D')$. As a consequence, while
    $D'$ is still in the stack $A'$, the exit condition of backjumping
    is not met. Moreover, once $D'$ is popped, the bound in
    $CS \setminus \{B_{top}\}$ with the highest height, which has
    decision level $l$, satisfies that there are no decisions above it
    in $A'$. Hence at this step backjumping halts.

  \end{itemize}

  Altogether, we conclude that $D'$ is the last bound that is popped
  from $A'$ in backjumping. And since for any bound
  $B \in CS \setminus \{B_{top}\}$ we have that
  $\height_{A}(B) < \height_{A}(D')$, it must be that
  $CS \setminus \{B_{top}\} \subseteq N$.

  To complete the proof of property \ref{def:valid:reason-set} in the
  definition of validity, we also have to see that
  $S \cup (CS \setminus \{B_{top}\}) \models \no{B_{top}}$, or
  equivalently, that
  $S \cup (CS \setminus \{B_{top}\}) \cup \{ B_{top} \} = S\cup CS$ is
  infeasible. But this is true, since it is an invariant of conflict
  analysis.


  \warning{Lastly, as regards property \ref{def:valid:reason-constraint} in
  the definition of validity, there is nothing to be proved. This
  concludes the proof of the theorem.}\comment{\qed}

\end{proof}

\bigskip

\begin{proof}[Proof of Theorem \ref{valid-cut-based}]

  To start with, following the same argument as in the proof of
  Theorem \ref{valid-resolution-based}, it is straightforward to see
  that the algorithm indeed terminates. Note that, as regards
  termination, the only difference between the procedure here and that
  in Section \ref{set-based} is that now \textsf{Early Backjump}
  allows exiting the loop earlier.

  Let us show property \ref{def:valid:consequence} in the definition
  of validity, that is, that for any constraint $C \in T$ we have
  $S \models C$. For that it is enough to see that $S \models CC$ is
  an invariant of conflict analysis. Let us prove it by induction on
  the number of steps. At the beginning $CC \in S$ by construction,
  and therefore $S \models CC$. Now let us assume that the property
  holds, and let us see that it is preserved in the next iteration of
  conflict analysis. Let $B$ be the bound in $CS$ that is topmost in
  $A$, and $x$ its variable. Let us assume that $RC$ the reason
  constraint of $B$ is defined and that there exists a cut eliminating
  $x$ between $CC$ and $RC$, which we will denote by $CC'$ (otherwise,
  $CC$ does not change and there is nothing to prove). Then by the
  properties of the cut rule, $CC \land RC \models CC'$. Moreover
  $S \models RC$ by properties \ref{item:constraint-set-sols} and
  \ref{item:stack-reason-constraint} of Lemma \ref{lemma:invariants}.
  Thus, the induction hypothesis $S \models CC$ together with
  $S \models RC$ and $CC \land RC \models CC'$ imply that
  $S \models CC'$, which completes the proof of invariance. Also as a
  byproduct, this also proves property
  \ref{def:valid:reason-constraint} in the definition of validity.

  Let us show property \ref{def:valid:decomposition} in the definition
  of validity. We have to see that the assignment $A$ can be
  decomposed as $N\, D\, M$, where $N$ and $M$ are sequences of bounds
  and $D$ is a decision bound, and $A'$ is of the form $N\, B$, where
  $B$ is a fresh bound in $N$. If \textsf{Backjump} is applied, this
  holds following the same argument as in the proof of Theorem
  \ref{valid-resolution-based}. On the other hand, if \textsf{Early
    Backjump} is applied, let $D$ be the last bound that is popped,
  which by construction is a decision. So the assignment $A$ can be
  decomposed as $N\, D\, M$, where $N$ and $M$ are sequences of bounds
  and $D$ is a decision bound. Moreover, $A'$ is of the form $N\, B'$,
  where by definition $B'$ is a fresh bound in $N$. Hence property
  \ref{def:valid:decomposition} is also true in this case.

  Finally, regarding property \ref{def:valid:reason-set} in the
  definition of validity, again following the same argument as in the
  proof of Theorem \ref{valid-resolution-based}, it is straightforward
  to see that it holds for bounds that are pushed in
  \textsf{Backjump}. For bounds that are pushed in \textsf{Early
    Backjump}, let $D$ be the last bound that is popped, which as
  observed above is a decision. Hence the assignment $A$ can be
  decomposed as $N\, D\, M$. If $B'$ is the bound that is pushed and
  $CC$ and $RS'$ are its reason constraint and its reason set, then by
  definition $RS' \subseteq N$ and $CC \cup RS' \models B'$, which
  together with $S \models CC$ implies $S \cup RS' \models B'$.\comment{\qed}
\end{proof}

\end{document}